\pdfoutput=1

\documentclass[11pt]{article}

\usepackage{acl}

\usepackage{times}
\usepackage{latexsym}

\usepackage[T1]{fontenc}

\usepackage[utf8]{inputenc}

\usepackage{microtype}

\usepackage{inconsolata}
\usepackage{booktabs}
\usepackage{multirow}
\usepackage{tabularx}
\usepackage{tcolorbox}
\usepackage[inline]{enumitem}
\newcolumntype{Y}{>{\centering\arraybackslash}X}
\newcommand{\modelname}{VividMed}

\usepackage{amsmath}
\usepackage{amssymb}
\DeclareMathOperator*{\argmin}{arg\,min}
\usepackage{nicefrac}
\usepackage{graphicx}
\usepackage{subcaption}
\usepackage{cleveref}
\usepackage{authblk}

\title{\modelname{}: Vision Language Model with Versatile Visual Grounding for Medicine}

\author{Lingxiao Luo\textsuperscript{*}}
\author{Bingda Tang\textsuperscript{*}}
\author{Xuanzhong Chen}
\author{Rong Han}
\author{Ting Chen\textsuperscript{\textdagger}}
\affil{
    Tsinghua University \\
    \texttt{\{luolx24,tbd21,cxz23,hanr21\}@mails.tsinghua.edu.cn}\\
    \texttt{tingchen@tsinghua.edu.cn}
}

\begin{document}
\maketitle
\renewcommand*{\thefootnote}{\fnsymbol{footnote}}
\footnotetext[1]{Equal contribution.}
\footnotetext[2]{Corresponding author.}
\renewcommand*{\thefootnote}{\arabic{footnote}}

\begin{abstract}
Recent advancements in Vision Language Models (VLMs) have demonstrated remarkable promise in generating visually grounded responses.
However, their application in the medical domain is hindered by unique challenges.
For instance, most VLMs rely on a single method of visual grounding, whereas complex medical tasks demand more versatile approaches.
Additionally, while most VLMs process only 2D images, a large portion of medical images are 3D.
The lack of medical data further compounds these obstacles.
To address these challenges, we present \textbf{\modelname{}}, a vision language model with versatile visual grounding for medicine.
Our model supports generating both semantic segmentation masks and instance-level bounding boxes, and accommodates various imaging modalities, including both 2D and 3D data.
We design a three-stage training procedure and an automatic data synthesis pipeline based on open datasets and models.
Besides visual grounding tasks, \modelname{} also excels in other common downstream tasks, including Visual Question Answering (VQA) and report generation.
Ablation studies empirically show that the integration of visual grounding ability leads to improved performance on these tasks.
Our code is publicly available at \url{https://github.com/function2-llx/MMMM}.
\end{abstract}

\section{Introduction}

Medical data encompasses a broad spectrum of modalities, such as medical images, radiology reports and genomics.
Synthesizing these diverse data is essential for building a holistic view of the health condition of a patient, enabling precision diagnostics and treatment planning.
The emergence of Large Multimodal Models (LMMs), particularly Vision Language Models (VLMs), has initiated a transformative paradigm shift in AI for medicine~\cite{li2023llavamed, bai2024m3d, wu2023radfm, wang2023r2gengpt, yang2024medgemini}.
LMMs overcome the limitations of task-specific specialist medical models confined to single input and output modalities, paving the way for the development of Generalist Medical AI (GMAI) models~\cite{Moor2023}.
These models are anticipated to handle dynamically specified tasks and incorporate versatile input and output modalities, and realize many unprecedented clinical use cases.

A prominent prospective use case of GMAI models is to draft grounded radiology reports~\cite{Moor2023}. 
Specifically, beyond providing textual reports for given radiology images, GMAI models could further ground the anatomical structures and abnormality findings mentioned by specific phrases in reports with localized visualizations, typically highlighting them with bounding boxes or segmentation masks.
Compared to plain textual reports, visually grounded reports possess significantly improved clinical utility by facilitating intuitive user interaction, effective interpretation of radiology images, and straightforward verification against harmful hallucinations.

Grounded report generation is rendered feasible by recent advancements in VLMs with visual grounding, which are capable of generating visually grounded detailed conversations~\cite{yang2023improved, Rasheed_2024_CVPR_GLaMM}.
Despite general-purpose visual grounding VLMs have demonstrated impressive performance, they still struggle to interpret medical images with accurate anatomical localization~\cite{zhou2024evaluating, wu2023can}. 
To address these limitations, we propose \modelname{}: \underline{Vi}sion Language Model with Versatile \underline{Vi}sual Groun\underline{d}ing for \underline{Med}icine, which supports diverse downstream tasks and accommodates both 2D and 3D imaging modalities. \modelname{} implements visual grounding via prompting localization modules based on the Segment Anything Model (SAM)~\cite{sam2023iccv} with the hidden embeddings of the additional special tokens in the VLM's output, which are then decoded into corresponding image regions.

Replicating the success of general-purpose VLMs with visual grounding capability in the medical domain is non-trivial, as several major challenges impede such efforts.
\begin{enumerate*}[label=(\textbf{\roman*})]
\item Medical images are highly heterogeneous, encompassing diverse imaging modalities. 
However, existing VLMs are predominantly developed for 2D natural images and are inherently inefficient at handling 3D medical images. To address this, we draw inspiration from previous works~\cite{SpadNets_arxiv_2024_Luo} and dynamically adjust the patch embeddings of the vision encoder. 
\item Existing grounding VLMs typically generate either segmentation masks or bounding boxes. However, both forms are essential for our tasks.
While some anatomical structures and abnormalities are best captured by segmentation masks, others are better delineated using bounding boxes as they are ill-suited for segmentation.
Therefore, we augment the localization module to generate both. 
\item Radiologists sometimes refer to multiple instances in a single phrase, therefore requiring instance segmentation and detection. For example, in a chest X-ray report, a radiologist might note ``multiple lung nodules are observed'', without specifying each nodule separately. We attend to this problem by adapting SAM to generate multiple outputs. 
\item The most significant obstacle is the scarcity of publicly available data.
Currently, there is no single dataset can support the development of grounded report generation. To tackle this challenge, we propose a three-stage training and automatic data annotation pipeline that make the best use of existing localization and report generation datasets to realize grounded report generation.
\end{enumerate*}

Experiments show that \modelname{} not only excels in previously unassailable visual grounding tasks, but also exhibits competitive performance on common downstream tasks such as visual question answering (VQA) and report generation.
Ablation studies also empirically show that the integration of visual grounding capability allows medical VLMs to achieve improved performance on other downstream tasks.
Our main contributions are summarized as follows:
\begin{itemize}
    \item We present \modelname{}, an exploratory attempt to equip medical VLMs with versatile visual grounding capabilities, paving the way for grounded report generation along with other visual grounding tasks.
    \item We design a three-stage training procedure for \modelname{} and an automatic data synthesis pipeline to tackle the scarcity of data, where all datasets and models involved are from the open domain.
    \item We conduct extensive experiments to validate the effectiveness of \modelname{} on various downstream tasks.
    The experimental results also show that integrating visual grounding ability to VLMs benefit downstream tasks.
\end{itemize}

\

\section{Related Works}

\subsection{Medical VLMs}

Building upon the remarkable success of general-purpose VLMs~\cite{dai2023instructblip, liu2024llavanext, wang2023cogvlm}, a multitude of medical VLMs have been developed for varying ranges of imaging modalities and downstream tasks~\cite{li2023llavamed, bai2024m3d, wu2023radfm, wang2023r2gengpt, yang2024medgemini, hyland2024maira1}.
Recently, the concept of GMAI~\cite{Moor2023} is gaining increasing attention for the promising clinical utility.
However, existing VLMs are mostly restricted to text generation tasks and 2D input images, limiting their real-world applications.
Our work aims to move a step forward towards GMAI, where we explore equipping medical VLMs with versatile visual grounding capabilities for both 2D and 3D imaging modalities.

\subsection{Visual Grounding VLMs}

A branch of existing visual grounding VLMs implement visual grounding by relying on the LLM to generate bounding box coordinates in literal texts~\cite{wang2023cogvlm,chen2023shikra,bai2023qwen} or discretized location tokens~\cite{peng2024grounding}. 
More recent approaches incorporate external pre-trained object detectors or segmentation models by prompting them with hidden embeddings from the VLM~\cite{pi-etal-2023-detgpt, you2024ferret, lai2023lisa, Rasheed_2024_CVPR_GLaMM, yang2023improved, zhang2024groundhog}.

Due to the aforementioned challenges, existing general-purpose VLMs with visual grounding do not generalize effectively to the medical domain, necessitating the development of domain-specific models.
M3D~\cite{bai2024m3d} implements referring expression segmentation by employing the promptable segmentation module, but it does not support grounded report generation and was trained solely on 3D images.
Concurrent to our work, MAIRA-2~\cite{bannur2024maira2groundedradiologyreport} offers promising results in grounded report generation.
However, it is tailored for grounded report generation on 2D Chest X-ray images and cannot be flexibly applied to other tasks and 3D images.
Moreover, it relies on a substantial amount of private data specifically annotated for grounded report generation, limiting the contribution to the open source community.

\section{Method}

\subsection{Task Formulation}

We begin by formulating the task of Vision Language models (VLMs) with visual grounding as considered in this work.
Similar to regular VLMs, such a model generates responses based on an input image and language instructions.
In addition to generating text, the model also identifies key phrases \(\{r_i\}_{i=1}^k\) within the generated text that refer to specific visual objects or regions of interest in the image.
For each identified phrase \(r_i\), the model maps it to corresponding localized representations, such as bounding boxes or segmentation masks, thereby making the responses visually grounded. 

The visual objects and regions of interest for visual grounding vary by application scenario.
In this work, we focus on developing VLMs for medical images that can visually ground anatomical structures and abnormalities, which are crucial in radiology.
In general, anatomical structures are grounded with segmentation masks, and abnormalities are grounded with bounding boxes.
These models can perform conventional tasks like medical VQA and report generation, as well as novel tasks requiring visual grounding such as grounded report generation, and target detection and localization.

\subsection{Model Architecture}

\modelname{}, our proposed vision-language model with visual grounding for medicine, is built upon a base VLM with an additional promptable localization module, as demonstrated in Figure~\ref{fig:model}.

\subsubsection{Base VLM}
\label{sec:vg-format}

\begin{figure*}
    \centering
    \includegraphics[width=.95\linewidth]{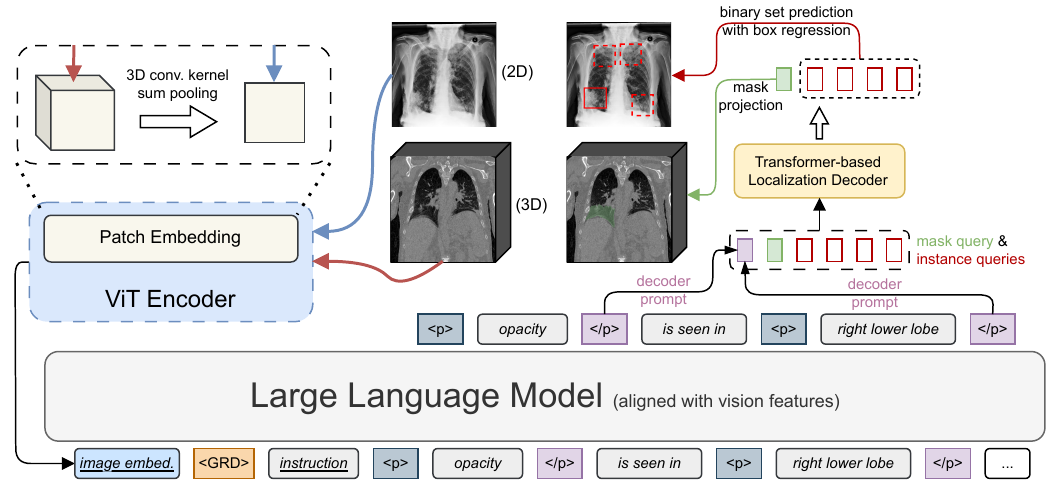}
    \caption{
        The architecture of \modelname{}, which is built upon a base VLM (left and lower) and a promptable localization module (upper right).
        The model identifies key phrases for grounding by enclosing them with bracket tokens, and the hidden states of the closed bracket token is used for prompting the localization module.
        The query tokens for both mask and instances are fed to the transformer-based localization decoder in parallel.
        The bounding boxes for negative instances are illustrated with dashed lines.
        The model accepts both 2D and 3D images as input by adaptively adjusting weights in the patch embedding layer.
        The vision encoder of the localization module is omitted for clarity.
    }
    \label{fig:model}
\end{figure*}

We adopt CogVLM~\cite{wang2023cogvlm} as our base VLM to generate responses given the input image and language instructions.
The detailed architecture of the base VLM is recapped in \Cref{sec:vlm}.
To enable the model to generate visually grounded responses, we draw inspiration from previous work~\cite{peng2024grounding, Rasheed_2024_CVPR_GLaMM, zhang2024groundhog} and 
 fintune the VLM to enclose the target phrases to be grounded with special bracket tokens, \texttt{<p>} and \texttt{</p>}, when generating their responses.
For instance, in the example shown in Figure~\ref{fig:model}, the model should generate the response \texttt{<p>opacity</p> is seen in <p>right lower lobe</p>}, where the anatomical structure ``right lower lobe'' and the abnormality ``opacity'' are enclosed within the bracket tokens.
Besides, two special tokens, \texttt{<grd>} and \texttt{<ngrd>}, are also introduced to indicate whether the model should perform visual grounding. We insert either of them at the beginning of a instruction, which helps the model adapt to training data with different granularity of available annotation, and serves as a switch for visual grounding during inference.

\subsubsection{Localization Module}

The architecture of the promptable localization module generally follows SAM~\cite{sam2023iccv}, which consists of a vision encoder and a transformer-based decoder.
For each phrase identified by the VLM to be grounded, we extract the last-layer hidden states of the corresponding closed bracket token \texttt{</p>} as its embedding. This embedding is then projected through an MLP to serve as the prompt for the decoder, which subsequently generates the corresponding bounding boxes or segmentation masks based on the encoded input image for each prompt.

We emphasize that enabling the SAM mask decoder to output bounding boxes is not as trivial as merely reducing from output segmentation masks or introducing an additional box prediction head.
The vanilla SAM mask decoder outputs only a single binary mask for each prompt, projected from a mask query token. Such behavior is insufficient to distinguish between different instances corresponding to the same phrase prompt\footnote{
    Strictly speaking, while the SAM mask decoder does predict multiple masks for a prompt to tackle ambiguity during training, these masks are not intended for distinguishing different instances, and only one of them will be selected.
}.
In particular, when annotations for different instances are available, they are typically in the form of bounding boxes.
Even the compromise of merging these bounding boxes into a single one will result in excessive information loss.

To address this challenge, we introduce a new branch to the decoder, in addition to the vanilla mask prediction branch, that predicts multiple different instances corresponding to the prompt, formulated as a binary set prediction task inspired by DETR-like methods~\cite{DeTR_ECCV_2020_Carion}.
Specifically, we introduce \(m\) additional instance query tokens to the decoder, where each token may correspond to a unique instance in the image or be dummy negative, indicating no correspondance to any instance. 
The number of tokens \(m\) is predefined to be larger than the number of different instances associated with a prompt in most medical images.
Let \(y = \{y_i\}_{i=1}^{m}\) and \(\hat{y} = \{\hat{y}_i\}_{i=1}^{m}\) denote the set of ground truth labels and predictions, respectively, with the ground truth padded to the size of \(m\) with dummy negative instances.
To compute the loss during training, each prediction \(i\) is first assigned a unique label \(\sigma(i)\), where the permutation \(\sigma \in S_m\) is determined by the following assignment objective:
\begin{equation}
    \argmin_{\sigma \in S_m}
    \sum_{i=1}^m L_{\mathrm{cost}}(\hat{y}_i, y_{\sigma(i)}).
    \label{eq:assignment}
\end{equation}
Given the cost function \(L_{\mathrm{cost}}\), the optimization objective \ref{eq:assignment} can be solved precisely in polynomial time using the Hungarian algorithm~\cite{Kuhn1955hungarian, Hungarian_Jacobi_1865}.

Let \(b_i\) and \(\hat{b}_i\) denote the bounding box coordinates for \(y_i\) and \(\hat{y}_i\), respectively;
\(c_i = 1\) if \(y_i\) is positive, and \(c_i = 0 \) otherwise;
and \(\hat{p}_i\) denotes the predicted probability of \(\hat{y}_i\) being positive.
The cost function \(L_{\mathrm{cost}}\) for each pair of \(\hat{y}_i\) and \(y_j\) is defined as a linear combination of a bounding box regression loss \(L_{\mathrm{box}}\) and a discrimination loss \(L_{\mathrm{disc}}\):
\begin{equation}
    L_{\mathrm{disc}}(\hat{p}_i, c_j) + \begin{cases}
        0 & c_j=0 \\
        L_{\mathrm{box}}(\hat{b}_i, b_j) & c_j=1
    \end{cases}.
\end{equation}
We choose \(\ell^1\) loss and GIoU loss~\cite{GIoU_CVPR_2019_Rezatofighi} for \(L_{\mathrm{box}}\), and focal loss~\cite{focal_loss_TPAMI_2020_Lin} for \(L_{\mathrm{disc}}\) following DINO~\cite{DINO_ICLR_2023_Zhang}.

The final loss for bounding boxes is the same as Eq.~\ref{eq:assignment}.
For segmentation masks, we use a combination of Dice loss and focal loss.

\subsubsection{Diverse Input Handling}

Most medical images consist of a stack of 2D image slices.
A common 2D image, such as an X-ray image, is a special case with a single slice.
To handle medical images with various numbers of slices, a direct approach would be to interpolate all images to a fixed size.
However, for 3D images, inter-slice interpolation can introduce unwanted artifacts, such as overlapping contents of adjacent slices.
Instead of interpolating input images, we dynamically adjust related model weights based on the number of slices of the input image, inspired by previous works on building universal backbones for medical images~\cite{SpadNets_arxiv_2024_Luo}. 
We provide a detailed discussion on this topic in \Cref{sec:discuss-input-handling}.

\paragraph{Patch Embedding}

For a ViT-based vision encoder, we set the maximum number of patches \(t_d\) and the base patch size \(P_d\) along the depth dimension, on which the image slices are stacked.
For an input image with \(D\) slices, the effective patch size \(P'_d\) is dynamically given as the closest valid patch size, with a closed form:
\begin{equation}
    \begin{cases}
        1 & D \le t_d \\
        2 \uparrow \mathrm{round}\left(\log_2{\frac{D}{t_d}}\right), & t_d < D \le t_d P_d \\
        P_d & D > t_d P_d
    \end{cases},
\end{equation}
where \(a \uparrow b = a^b\) and \(\mathrm{round}(x) = \lfloor x + \nicefrac12 \rfloor \).
The convolution kernel weight in the patch embedding layer is then reduced to the effective patch size through sum pooling, which make the output embeddings of different patch sizes commensurable.
During training, we sample \(\log_2 P'_d\) from a normal distribution \(N(\log_2(\nicefrac{D}{t_d}), 0.25)\) for augmentation.

\paragraph{Upsampling}

The decoder in the localization module involves the upsampling of feature maps when output segmentation masks.
The upsampling is achieved with a series of transposed convolution layers, each with a scale factor of 2. 
To preserve a consistent size with the input, if the depth of the feature map has already reached \(D\), then upsampling is disabled along the depth dimension.
This is implemented with reducing the transposed convolution kernel weight along the depth dimension with mean pooling.

\subsection{Model Training}

We design a three-stage training procedure for \modelname{}, where each stage involves different training tasks.
All stages are trained end-to-end with visual instruction-following training data, which are constructed using open datasets and models.
We sketch the training procedure in \Cref{sec:training-procedure} and detail each task involved in \Cref{sec:tasks}.

\subsubsection{Training Procedure}
\label{sec:training-procedure}

\paragraph{Stage 1: Visual Grounding Pre-training}

In the first stage, we pre-train the model's visual grounding ability with the task of target detection and localization.
Specifically, the model is instructed to determine whether given targets exist on the image, and list the target names along with their presence in the response.
The target names in responses are then visually grounded on the image.
We utilize open source medical image semantic segmentation and disease detection datasets to construct training data for this task.

\paragraph{Stage 2: Medical Visual Instruction Tuning}

This stage is dedicated to training the model's visual understanding and reasoning capabilities for medical images. The training data is constructed using hand-crafted prompt templates across several tasks, including visual question answering (VQA), image captioning, and report generation. Visual grounding is disabled during this stage.

\paragraph{Stage 3: Alignment}

In the third stage, we fine-tune the model to align both the visual grounding and medical image understanding abilities trained by previous stages to unleash the combined strengths.
To do this, we train the model with the grounded report generation task, where the model generates reports for input images and visually grounds key phrases on images.
We synthesize training data for this stage as described in \Cref{sec:grg-data}.

\subsubsection{Training Datasets}
\label{sec:tasks}



\paragraph{Visual Question Answering}

This task involves instructing the model with a question about an image, and the model should answer based on the image.
We construct three types of VQA data.
\begin{enumerate*}[label=(\roman*)]
    \item \textbf{Modality Recognition}:
    query the imaging modality of the image, such as X-ray, CT, and MRI.
    The modality information is available for most data, and we randomly include this task for 50\% of training samples.
    \item \textbf{Plane Recognition}:
    query the viewing plane of the chest X-ray image.
    We randomly include this task for 20\% of training samples from MIMIC-CXR.
    \item \textbf{Abnormality Recognition}:
    query if specific abnormalities are present on the input image.
    We randomly include this task for 20\% of training samples from VinDr-CXR, MIMIC-CXR, and CT-RATE, utilizing the associated abnormality labels.
\end{enumerate*}

\paragraph{Image Captioning}
This task involves the model predicting the caption of an image.
We adopt the ROCOv2~\cite{rückert2024rocov2} dataset for this task, which comprises 79,789 diverse radiographs with associated medical concepts and captions.
We address hallucination vulnerability in captions as described in \Cref{sec:data-cleaning}, and discard captions with overly low quality. After filtering, there are 59,958 image-caption pairs.

\paragraph{Report Generation}

This task requires the model to generate two key sections of a typical report: 
\begin{enumerate*}[label=(\roman*)]
    \item \textbf{Findings}:~\cite{johnson2019mimiccxr}:
    provides a detailed description of the observations from the imaging study, including the presence of any abnormalities and their anatomical locations.
    \item \textbf{Impression}:
    synthesizes these observations into a concise diagnostic summary.
\end{enumerate*}

We employ two large publicly available radiology report datasets, encompassing both 2D and 3D images:
\begin{enumerate}[label=(\roman*)]
    \item \textbf{MIMIC-CXR}~\cite{johnson2019mimiccxr}:
    A chest X-ray dataset containing 377,110 images corresponding to 227,835 radiographic studies, each study accompanied by labels and a report.
    In this work, we use its JPEG format version~\cite{johnson2019mimiccxrjpg}.
    \item \textbf{CT-RATE}~\cite{hamamci2024ctrate}: A 3D medical imaging dataset consisting of 25,692 chest CT volumes paired with labels and reports.
\end{enumerate}

For both datasets, we use official data splits and address hallucination vulnerability in the reports as described in ~\Cref{sec:data-cleaning}. We also discard reports lacking ``Findings'' or ``Impression'' sections.
For MIMIC-CXR, we filter the training set for balance between studies with and without findings.
In line with prior studies~\cite{yang2024medgemini, hyland2024maira1}, we only use frontal chest X-ray images to generate report, which visualize the anatomy most clearly.
After filtering, there are 121,953 training and 1,587 testing image-report pairs in MIMIC-CXR and 24,086 training and 1,560 testing image-report pairs in CT-RATE. 

\paragraph{Grounded Report Generation}

\begin{table}[b]
    \begin{tabular}{c|cc}
    \toprule
    \textbf{Dataset}       & MIMIC-CXR & CT-RATE \\
    \midrule
    \#tags        & 435396    & 346650  \\
    \#boxes/masks & 33114     & 96620  \\
    \bottomrule
    \end{tabular}
    \caption{
        Statistics for resulting grounded reports generated by our pipeline. Note that the number of boxes or masks are significantly smaller than tags due to many classes are unsupported by the pre-trained detection or segmentation module.
    }
    \label{table:gr-stat}
\end{table}

We construct training data for the task with an automatic pipeline using open datasets and models, the details are described in \Cref{sec:grg-data}.
The statistics for resulting grounded reports are presented in Table~\ref{table:gr-stat}.

\subsubsection{Grounded Reports Construction}
\label{sec:grg-data}

We design a automatic pipeline to construct training data for the task of grounded report generation.
The pipeline is applied to both MIMIC-CXR and CT-RATE datasets.

\paragraph{Key Phrases Identification}

First, we instruct the pre-trained LLM of Meta Llama 3 to identify key phrases in the report text that correspond to anatomical structures or abnormality findings on images~(Figure~\ref{fig:tag-prompt}).
In our experiments, we find that the fully open-vocabulary manner for this step results in inferior results.
Therefore, we maintain a taxonomy of common targets of human body and instruct the LLM to focus on targets within it.
As a result, key phrases along with their standardized names are extracted from the reports.

\paragraph{Positive Targets Filtering}

We find that LLM tends to wrongly identify targets that are stated as absent in the image, such as ``No pleural effusion or pneumothorax is observed''.
Therefore, we introduce an intermediate step by instructing the LLM to filter only positive targets from the output of the last step (Figure~\ref{fig:filter-prompt}).

\paragraph{Localized Annotations Generation}

Finally, we utilize pre-trained models to generate localized annotations for extracted phrases.
For abnormality targets, we train a detection model of DINO with EVA-02 backbone~\cite{FANG2024105171_EVA02} ourselves, utilizing the VinDr-CXR dataset.
For anatomical structures, we simply utilize the pre-trained SAT-Pro~\cite{zhao2024modelrulealluniversal} as it demonstrates robust out-of-box segmentation performance.

\begin{table*}[h]
\centering
\fontsize{8.5pt}{\baselineskip}\selectfont
\begin{tabularx}{\textwidth}{@{}c|YcY|YcY|YcY@{}}
\toprule
                     & \multicolumn{3}{c|}{VQA-RAD} & \multicolumn{3}{c|}{SLAKE}   & \multicolumn{3}{c}{VQA-Med} \\ 
Model                & BLEU-1 & ROUGE-1 & Accuracy & BLEU-1 & ROUGE-1 & Accuracy & BLEU-1 & ROUGE-1 & Accuracy \\ \midrule
InstructBLIP         & 0.368   & 0.392    &     0.428     & 0.510   & 0.551    &     0.558     & 0.166   & 0.205    &     0.222     \\
LLaVA 1.6 (13B) & 0.526   & 0.540    &    0.558      & 0.818   & 0.822    &     0.828     & 0.619   & 0.630    &    0.614      \\
CogVLM               & \textbf{0.545}   & \textbf{0.559}    &     \textbf{0.568}     & \underline{0.840}   & \underline{0.843}    &      \underline{0.832}    & \underline{0.621}   & \underline{0.631}    &      \underline{0.621}    \\
LLaVA-Med 1.5           &     0.491   &    0.503     &     0.529     &    0.579    &     0.581    &     0.559     &   0.398     &     0.400    &     0.391     \\
M3D                  &    0.471    &    0.481     &     0.497     &    0.557    &    0.570     &    0.544      &    0.272    &    0.270     &     0.263     \\
RadFM                &   \textcolor{lightgray}{0.541}   &   \textcolor{lightgray}{0.557}   &    \textcolor{lightgray}{0.588}      &     0.784   &     0.789    &     0.771     &    \textcolor{lightgray}{0.519}    &    \textcolor{lightgray}{0.536}     &     \textcolor{lightgray}{0.543}     \\ \midrule
\textbf{\modelname{}}~\textsubscript{w/o VG}           &   0.519     &     0.533    &   0.566       &    0.878    &    0.882    &     0.869     &    0.623   &     0.633     &    0.619      \\ 
\textbf{\modelname{}}      &    \underline{0.542}    &     \underline{0.558}    &    \textbf{0.568}      &    \textbf{0.880}    &    \textbf{0.885}     &   \textbf{0.873}   &     \textbf{0.636}   &     \textbf{0.648}    &    \textbf{0.637}     \\ 

\bottomrule
\end{tabularx}
\caption{
    Evaluation results of visual question answering. Accuracy is evaluated with Llama 3 70B.
    We notice that RadFM is trained on the \href{https://medpix.nlm.nih.gov}{MedPix\textsuperscript{\textregistered}} database, which is the source for both VQA-RAD and VQA-Med. Therefore, we exclude RadFM from comparison on both datasets.
    }
\label{tab:vqa}
\end{table*}

\begin{table*}[h]
    \small
    \centering
    \fontsize{8.5pt}{\baselineskip}\selectfont
    \begin{tabularx}{\textwidth}{c  c | Y  Y  Y  Y  Y}
    \toprule
    Dataset                    & Metric         &    R2GenGPT     & M3D                  & RadFM        &    \textbf{\modelname{}}~\textsubscript{w/o VG}   & \textbf{\modelname{}}   \\ \midrule
    \multirow{14}{*}{MIMIC-CXR} & BLEU-4              &        0.093     &    0.049     &      0.071     &      \textbf{0.122}     &           0.120           \\
                               & ROUGE-L             &        0.267     &    0.200     &      0.253      &     \textbf{0.310}     &           0.306           \\
                               & METEOR              &     0.310     &         0.241   &      0.283     &      0.361     &         \textbf{0.364}             \\
                               & Macro CheXpert F1 14        &     0.295      &     0.115      &    0.165    &       0.346       &      \textbf{0.370}                \\
                               & Micro CheXpert F1 14     &     0.440      &      0.176     &      0.268     &     0.507      &      \textbf{0.529}                \\
                               & Macro CheXpert F1 5        &     0.453      &     0.193      &    0.279    &       0.494       &      \textbf{0.512}                \\
                               & Micro CheXpert F1 5     &     0.522      &      0.234     &      0.361     &     0.579      &      \textbf{0.598}                \\
                               & Macro CheXpert FNR 14 & 0.152 & 0.199 & 0.177 & 0.138 & \textbf{0.133} \\
                               & Micro CheXpert FNR 14 & 0.146 & 0.195 & 0.178 & 0.131 & \textbf{0.124} \\
                               & Macro CheXpert FNR 5 & 0.218 & 0.293 & 0.257 & 0.190 & \textbf{0.181} \\
                               & Micro CheXpert FNR 5 & 0.209 & 0.285 & 0.251 & 0.186 & \textbf{0.175} \\
                               & CheXbert Similarity &       0.393       &    0.251    &      0.299     &      0.444     &        \textbf{0.445}              \\
                               & RadGraph F1         &       0.240      &     0.169    &     0.182      &      \textbf{0.278}     &       \textbf{0.278}               \\
                               & RadCliQ v1 ($\downarrow$)            &     0.272       &     0.504     &     0.423     &       \textbf{0.142}     &   \textbf{0.142}                   \\ \midrule
    \multirow{7}{*}{CT-RATE}   & BLEU-4              &    -     &      0.193       &     0.226      &       0.240    &         \textbf{0.245}             \\
                               & ROUGE-L             &     -     &     0.327       &     0.352     &     0.369       &         \textbf{0.373}             \\
                               & METEOR              &      -      &     0.343     &     0.402     &      0.418      &    \textbf{0.419}                  \\
                               & Macro RadBERT F1         &     -       &     0.114     &   0.112        &      0.264     &      \textbf{0.312}                \\
                               & Micro RadBERT F1   &      -      &     0.182      &     0.215      &     0.375      &   \textbf{0.395}           \\ 
                               & Macro RadBERT FNR & - & 0.192 & 0.184 & 0.160 & \textbf{0.156} \\
                               & Micro RadBERT FNR & - & 0.183 & 0.176 & 0.152 & \textbf{0.149} \\
    \bottomrule
    \end{tabularx}
    \caption{
        Evaluation results of report generation on MIMIC-CXR and CT-RATE test sets.
        Note that R2GenGPT can only handle 2D images and is not evaluated on CT-RATE, where the images are 3D CT scans.
    }
    \label{tab:report}
\end{table*}

\section{Experiments}





\begin{figure*}[htbp]
    \centering
    \includegraphics[width=\linewidth]{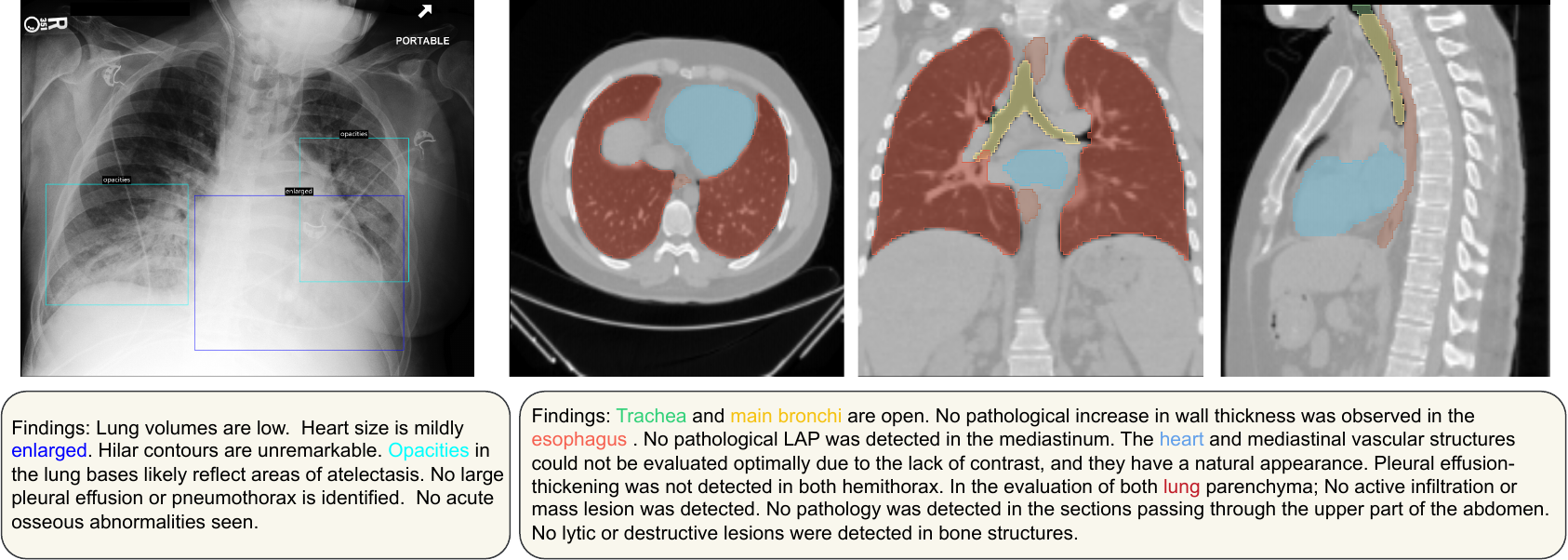}
    \caption{
        Selected qualitative results for grounded report generation, zoom in for better view.
        Impressions are omitted for clarity.
    }
    \label{fig:grg-examples}
\end{figure*}

\subsection{Target Detection and Localization}

\paragraph{Datasets}
We use the validation split of TotalSegmentator~(TS)~\cite{totalsegmentator} for segmentaiton mask generation and the test split of VinDr-CXR for bounding boxes generation.
\paragraph{Settings}
In this task, we evaluate the model's localization ability for given targets.
Specifically, the model directly processes class names in visual grounding format, as described in \Cref{sec:vg-format}, and the hidden states of the \texttt{</p>} is used to prompt the localization module and obtain the final results.
\paragraph{Metrics}
For segmentation, we compute Dice coefficient and \(\ell^1\) distance.
\paragraph{Results}
The results are as follows:
\begin{center}
    \small
    \begin{tabular}{l|ccc}
         & Dice (\%) & Mean \(\ell^1\) & Mean GIoU \\
        \midrule
        nnU-Net & 84.0 & - & - \\
        \modelname{} & 70.3      & 0.121   & 1.43 \\
    \end{tabular}
\end{center}
The performance of the nnU-Net~\cite{Isensee_2020_nnunet} is taken from official TS results.
It is expected that results are inferior than models specific for segmentation or detection, due to the limited training scale of our model and the exhaustive task prior incorporation by task-specific models.
We leave the improvement of localization quality to future works.


\subsection{Visual Question Answering}

\paragraph{Datasets}

We adopt three widely used VQA datasets for evaluation:
\begin{enumerate}[label=(\roman*)]
    \item \textbf{VQA-RAD}~\cite{lau2018vqarad}: A radiology VQA dataset comprising 315 images and 2,248 QA pairs.
    \item \textbf{SLAKE}~\cite{liu2021slake}: A bilingual (English and Chinese) medical VQA dataset. We only keep the English portion in our experiments, resulting in 641 images and 7,033 QA pairs.
    \item \textbf{VQA-Med}~\cite{VQA-Med2019}: A medical VQA dataset consisting of 4,200 images and 13,792 QA pairs.
\end{enumerate}

\paragraph{Settings}
We compare \modelname{} with several popular general-purpose and domain-specific models~\cite{dai2023instructblip, liu2024llavanext, wang2023cogvlm, li2023llavamed, bai2024m3d, wu2023radfm}.
All models are fine-tuned on each dataset for evaluation.
During training, we combine all available QA pairs for the same image into a multi-round conversation for better efficiency.

\paragraph{Metrics}
We employ BLEU-1~\cite{papineni2002bleu} and ROUGE-1~\cite{lin2004rouge} as evaluation metrics. As high-quality answers may not lexically match the reference ones, especially in medical contexts, we additionally utilize Llama 3 70B to evaluate accuracy.

\paragraph{Results}
The evaluation results are presented in Table.~\ref{tab:vqa}.
\modelname{} shows non-trivial general improvement over fine-tuned CogVLM and outperforms all other baselines.
Specifically, \modelname{} improves the answer accuracy by 4.1\% for SLAKE and 1.6\% for VQA-Med.
We also find that general-purpose VLMs like CogVLM and LLaVA 1.6 could also achieve promising results after fine-tuning on medical data, and we suggest that an effective way towards GMAI could still be starting from strong general-purpose foundation models and incorporating domain-specific data and designs for medical purposes.

\subsection{Report Generation}

\paragraph{Datasets}

The test sets of both MIMIC-CXR and CT-RATE are used for evaluation.

\paragraph{Settings}

Due to the complexity of report generation, we only focus on baselines that have undergone extensive training for this task~\cite{wang2023r2gengpt, bai2024m3d, wu2023radfm}.
For fair comparison, we further fine-tune all baselines (except for R2GenGPT, which is specialized for the MIMIC-CXR and OpenI datasets) on training sets to ensure output alignment.

\paragraph{Metrics}

Following common practices, we employ several common n-gram-based lexical metrics: BLEU-4, ROUGE-L and METEOR~\cite{banerjee2005meteor}.
We also evaluate the generated reports through the lens of clinical metrics, including CheXpert F1 and FNR, CheXbert vector similarity, RadGraph F1 and FNR, RadCliQ v1 and RadBERT F1. The details of the clinical metrics are given in \Cref{sec:clinical-report-metrics}.

\paragraph{Results}

The evaluation results are shown in \autoref{tab:report}.
\modelname{} outperforms all other baselines by a large margin on both datasets.
Given the higher BLEU-4, ROUGE-L and METEOR metrics on both dataset of \modelname{}, it could generate more coherent reports within the context of radiology, facilitating accurate interpretation of the generated reports for clinicians.
\modelname{} is also shown with stronger ability of abnormality recognition, where it improves macro CheXpert F1 by 8.5\% and macro RadBERT F1 by 4.8\%, and has FNRs consistently lower than other baselines.
Notably, \modelname{} is evaluated on both datasets directly without further fine-tuning on each dataset, highlighting its capability to effectively handle both 2D and 3D data simultaneously.


\subsection{Grounded Report Generation}

We evaluate \modelname{} on the corresponding test sets of both MIMIC-CXR and CT-RATE.
The selected qualitative results are shown in \Cref{fig:grg-examples}.
After alignment, \modelname{} is able to generate accurate report while also grounds key phrases on images, significantly enhancing interpretation procedure for medical images.
More results and analyses can be found in \Cref{sec:grg-qualitative}.

\paragraph{Error Analysis}

We conduct a qualitative review for grounded reports generated by \modelname{} on 12 selected cases from MIMIC-CXR test set, following the methodology of MAIRA-2~\cite{bannur2024maira2groundedradiologyreport}.
Among 91 generated sentences, 82 can be accepted as-is, 9 are wrongly stated and need major correction.
6 critical omissions are determined.
Overall, 8 reports (67\%) required at most one correction.
Among 17 correctly identified findings, 16 (94\%) of them are accurately visually grounded. 
Qualitative examples also show that when findings are incorrectly reported, visual grounding gives obvious outlier results.

\subsection{Ablation Studies}

We remove the visual grounding tasks from our training procedure to explore their impact on downstream tasks.
Results presented in \autoref{tab:vqa} and \autoref{tab:report} demonstrate consistent performance degradation, showing that the integration of visual grounding ability leads to improved performance on other downstream tasks.

Contrary to our findings, MAIRA-2~\cite{bannur2024maira2groundedradiologyreport} reports that integrating the grounded report generation task does not affect the report generation performance of the model.
We hypothesize that one reason behind this is that MAIRA-2 implements visual grounding with tokenized bounding box coordinates, which are still generated in the way of causal language modeling and the localized information is not effectively utilized.
On the other hand, our model incorporates a pixel- or voxel-level localization module and is end-to-end trained, which benefits both tasks.

\section{Conclusion}

In this paper, we present \modelname{} as a pioneering step towards vision-language models with versatile visual grounding for medical images.
Through its novel architecture, grounded data annotation pipeline, and the three-stage training procedure, \modelname{} exhibits superior performance on various downstream tasks, and realizes the visual grounding tasks especially the grounded report generation on MIMIC-CXR and CT-RATE datasets.
Our empirical results show that the integration of visual grounding capabilities boosts the performance of medical VLMs on other downstream tasks as well.
We believe our work has established a robust baseline in this field, and hope that future research may focus on improving performance further, as well as integrating into reliable clinical applications that benefit patients.

\section*{Limitations}

During our experiments, we observe that there is still room for improvement in downstream tasks through more careful hyperparameters tuning and more computational resources.
The clinical utility of our model can be further enhanced by allowing more flexible interaction.
In addition, we believe the incorporation of instance-level localization into visual grounding VLMs can be implemented with more recent advanced techniques, such as deriving from recent open-set object detection techniques, as well as function calling to external localization modules.
Furthermore, due to the limited available data, our model does not fully unleash the promising potentials of grounded report generation and struggles generalizing beyond chest X-ray and CT images on this task.
The absence of well-established evaluation metrics and benchmarks also poses a challenge in assessing the performance.




\section*{Ethics Statement}

All data involved in our study are sourced from publicly available, de-identified datasets.
Our model and data is not intended for real-world clinical usage.
Despite impressive performance compared to baselines, the generated reports still suffer from inaccuracies and require human review if applied in practice.
We recognize that while automated tools can enhance efficiency, the expertise of healthcare professionals remain indispensable for clinical practice in the foreseeable future.

\section*{Acknowledgement}

This study was supported by grants from the National Key R\&D Program of China (2024YFF1207100, 2024YFF1207103, 2022YFC2703100, 2022YFC2703105), Guoqiang Institute of Tsinghua University, and Beijing National Research Center for Information Science and Technology (BNRist). The funders had no roles in study design, data collection and analysis, publication decisions, or manuscript preparation.


\bibliography{custom}

\begin{thebibliography}{60}
\providecommand{\natexlab}[1]{#1}

\bibitem[{Ansel et~al.(2024)Ansel, Yang, He, Gimelshein, Jain, Voznesensky, Bao, Bell, Berard, Burovski, Chauhan, Chourdia, Constable, Desmaison, DeVito, Ellison, Feng, Gong, Gschwind, Hirsh, Huang, Kalambarkar, Kirsch, Lazos, Lezcano, Liang, Liang, Lu, Luk, Maher, Pan, Puhrsch, Reso, Saroufim, Siraichi, Suk, Zhang, Suo, Tillet, Zhao, Wang, Zhou, Zou, Wang, Mathews, Wen, Chanan, Wu, and Chintala}]{pytorch2}
Jason Ansel, Edward Yang, Horace He, Natalia Gimelshein, Animesh Jain, Michael Voznesensky, Bin Bao, Peter Bell, David Berard, Evgeni Burovski, Geeta Chauhan, Anjali Chourdia, Will Constable, Alban Desmaison, Zachary DeVito, Elias Ellison, Will Feng, Jiong Gong, Michael Gschwind, Brian Hirsh, Sherlock Huang, Kshiteej Kalambarkar, Laurent Kirsch, Michael Lazos, Mario Lezcano, Yanbo Liang, Jason Liang, Yinghai Lu, C.~K. Luk, Bert Maher, Yunjie Pan, Christian Puhrsch, Matthias Reso, Mark Saroufim, Marcos~Yukio Siraichi, Helen Suk, Shunting Zhang, Michael Suo, Phil Tillet, Xu~Zhao, Eikan Wang, Keren Zhou, Richard Zou, Xiaodong Wang, Ajit Mathews, William Wen, Gregory Chanan, Peng Wu, and Soumith Chintala. 2024.
\newblock \href {https://doi.org/10.1145/3620665.3640366} {Pytorch 2: Faster machine learning through dynamic python bytecode transformation and graph compilation}.
\newblock In \emph{Proceedings of the 29th ACM International Conference on Architectural Support for Programming Languages and Operating Systems, Volume 2}, ASPLOS '24, page 929–947, New York, NY, USA. Association for Computing Machinery.

\bibitem[{Bai et~al.(2024)Bai, Du, Huang, Meng, and Zhao}]{bai2024m3d}
Fan Bai, Yuxin Du, Tiejun Huang, Max Q.~H. Meng, and Bo~Zhao. 2024.
\newblock \href {https://arxiv.org/abs/2404.00578} {M3d: Advancing 3d medical image analysis with multi-modal large language models}.
\newblock \emph{Preprint}, arXiv:2404.00578.

\bibitem[{Bai et~al.(2023)Bai, Bai, Yang, Wang, Tan, Wang, Lin, Zhou, and Zhou}]{bai2023qwen}
Jinze Bai, Shuai Bai, Shusheng Yang, Shijie Wang, Sinan Tan, Peng Wang, Junyang Lin, Chang Zhou, and Jingren Zhou. 2023.
\newblock Qwen-vl: A frontier large vision-language model with versatile abilities.
\newblock \emph{arXiv preprint arXiv:2308.12966}.

\bibitem[{Banerjee and Lavie(2005)}]{banerjee2005meteor}
Satanjeev Banerjee and Alon Lavie. 2005.
\newblock Meteor: An automatic metric for mt evaluation with improved correlation with human judgments.
\newblock In \emph{Proceedings of the acl workshop on intrinsic and extrinsic evaluation measures for machine translation and/or summarization}, pages 65--72.

\bibitem[{Bannur et~al.(2024)Bannur, Bouzid, Castro, Schwaighofer, Bond-Taylor, Ilse, Pérez-García, Salvatelli, Sharma, Meissen, Ranjit, Srivastav, Gong, Falck, Oktay, Thieme, Lungren, Wetscherek, Alvarez-Valle, and Hyland}]{bannur2024maira2groundedradiologyreport}
Shruthi Bannur, Kenza Bouzid, Daniel~C. Castro, Anton Schwaighofer, Sam Bond-Taylor, Maximilian Ilse, Fernando Pérez-García, Valentina Salvatelli, Harshita Sharma, Felix Meissen, Mercy Ranjit, Shaury Srivastav, Julia Gong, Fabian Falck, Ozan Oktay, Anja Thieme, Matthew~P. Lungren, Maria~Teodora Wetscherek, Javier Alvarez-Valle, and Stephanie~L. Hyland. 2024.
\newblock \href {https://arxiv.org/abs/2406.04449} {Maira-2: Grounded radiology report generation}.
\newblock \emph{Preprint}, arXiv:2406.04449.

\bibitem[{{Ben Abacha} et~al.(2019){Ben Abacha}, Hasan, Datla, Liu, Demner-Fushman, and M\"uller}]{VQA-Med2019}
Asma {Ben Abacha}, Sadid~A. Hasan, Vivek~V. Datla, Joey Liu, Dina Demner-Fushman, and Henning M\"uller. 2019.
\newblock \href {https://ceur-ws.org/Vol-2380/paper\_272.pdf} {Vqa-med: Overview of the medical visual question answering task at imageclef 2019}.
\newblock In \emph{Working Notes of {CLEF} 2019}, volume 2380 of \emph{{CEUR} Workshop Proceedings}, Lugano, Switzerland. CEUR-WS.org.

\bibitem[{Cardoso et~al.(2022)Cardoso, Li, Brown, Ma, Kerfoot, Wang, Murrey, Myronenko, Zhao, Yang, Nath, He, Xu, Hatamizadeh, Myronenko, Zhu, Liu, Zheng, Tang, Yang, Zephyr, Hashemian, Alle, Darestani, Budd, Modat, Vercauteren, Wang, Li, Hu, Fu, Gorman, Johnson, Genereaux, Erdal, Gupta, Diaz-Pinto, Dourson, Maier-Hein, Jaeger, Baumgartner, Kalpathy-Cramer, Flores, Kirby, Cooper, Roth, Xu, Bericat, Floca, Zhou, Shuaib, Farahani, Maier-Hein, Aylward, Dogra, Ourselin, and Feng}]{cardoso2022monaiopensourceframeworkdeep}
M.~Jorge Cardoso, Wenqi Li, Richard Brown, Nic Ma, Eric Kerfoot, Yiheng Wang, Benjamin Murrey, Andriy Myronenko, Can Zhao, Dong Yang, Vishwesh Nath, Yufan He, Ziyue Xu, Ali Hatamizadeh, Andriy Myronenko, Wentao Zhu, Yun Liu, Mingxin Zheng, Yucheng Tang, Isaac Yang, Michael Zephyr, Behrooz Hashemian, Sachidanand Alle, Mohammad~Zalbagi Darestani, Charlie Budd, Marc Modat, Tom Vercauteren, Guotai Wang, Yiwen Li, Yipeng Hu, Yunguan Fu, Benjamin Gorman, Hans Johnson, Brad Genereaux, Barbaros~S. Erdal, Vikash Gupta, Andres Diaz-Pinto, Andre Dourson, Lena Maier-Hein, Paul~F. Jaeger, Michael Baumgartner, Jayashree Kalpathy-Cramer, Mona Flores, Justin Kirby, Lee A.~D. Cooper, Holger~R. Roth, Daguang Xu, David Bericat, Ralf Floca, S.~Kevin Zhou, Haris Shuaib, Keyvan Farahani, Klaus~H. Maier-Hein, Stephen Aylward, Prerna Dogra, Sebastien Ourselin, and Andrew Feng. 2022.
\newblock \href {https://arxiv.org/abs/2211.02701} {Monai: An open-source framework for deep learning in healthcare}.
\newblock \emph{Preprint}, arXiv:2211.02701.

\bibitem[{Carion et~al.(2020)Carion, Massa, Synnaeve, Usunier, Kirillov, and Zagoruyko}]{DeTR_ECCV_2020_Carion}
Nicolas Carion, Francisco Massa, Gabriel Synnaeve, Nicolas Usunier, Alexander Kirillov, and Sergey Zagoruyko. 2020.
\newblock End-to-end object detection with transformers.
\newblock In \emph{European conference on computer vision}, pages 213--229. Springer.

\bibitem[{Chen et~al.(2023)Chen, Zhang, Zeng, Zhang, Zhu, and Zhao}]{chen2023shikra}
Keqin Chen, Zhao Zhang, Weili Zeng, Richong Zhang, Feng Zhu, and Rui Zhao. 2023.
\newblock Shikra: Unleashing multimodal llm's referential dialogue magic.
\newblock \emph{arXiv preprint arXiv:2306.15195}.

\bibitem[{Chiang et~al.(2023)Chiang, Li, Lin, Sheng, Wu, Zhang, Zheng, Zhuang, Zhuang, Gonzalez, Stoica, and Xing}]{vicuna2023}
Wei-Lin Chiang, Zhuohan Li, Zi~Lin, Ying Sheng, Zhanghao Wu, Hao Zhang, Lianmin Zheng, Siyuan Zhuang, Yonghao Zhuang, Joseph~E. Gonzalez, Ion Stoica, and Eric~P. Xing. 2023.
\newblock \href {https://lmsys.org/blog/2023-03-30-vicuna/} {Vicuna: An open-source chatbot impressing gpt-4 with 90\%* chatgpt quality}.

\bibitem[{Dai et~al.(2023)Dai, Li, Li, Tiong, Zhao, Wang, Li, Fung, and Hoi}]{dai2023instructblip}
Wenliang Dai, Junnan Li, Dongxu Li, Anthony Tiong, Junqi Zhao, Weisheng Wang, Boyang Li, Pascale Fung, and Steven Hoi. 2023.
\newblock \href {https://openreview.net/forum?id=vvoWPYqZJA} {Instruct{BLIP}: Towards general-purpose vision-language models with instruction tuning}.
\newblock In \emph{Thirty-seventh Conference on Neural Information Processing Systems}.

\bibitem[{Dao(2024)}]{dao2024flashattention2}
Tri Dao. 2024.
\newblock \href {https://openreview.net/forum?id=mZn2Xyh9Ec} {Flashattention-2: Faster attention with better parallelism and work partitioning}.
\newblock In \emph{The Twelfth International Conference on Learning Representations}.

\bibitem[{Dao et~al.(2022)Dao, Fu, Ermon, Rudra, and R\'{e}}]{flashattn}
Tri Dao, Dan Fu, Stefano Ermon, Atri Rudra, and Christopher R\'{e}. 2022.
\newblock \href {https://proceedings.neurips.cc/paper_files/paper/2022/file/67d57c32e20fd0a7a302cb81d36e40d5-Paper-Conference.pdf} {Flashattention: Fast and memory-efficient exact attention with io-awareness}.
\newblock In \emph{Advances in Neural Information Processing Systems}, volume~35, pages 16344--16359. Curran Associates, Inc.

\bibitem[{Dubey et~al.(2024)Dubey, Jauhri, Pandey, Kadian, Al-Dahle, Letman et~al.}]{dubey2024llama3herdmodels}
Abhimanyu Dubey, Abhinav Jauhri, Abhinav Pandey, Abhishek Kadian, Ahmad Al-Dahle, Aiesha Letman, et~al. 2024.
\newblock \href {https://arxiv.org/abs/2407.21783} {The llama 3 herd of models}.
\newblock \emph{Preprint}, arXiv:2407.21783.

\bibitem[{Fang et~al.(2024)Fang, Sun, Wang, Huang, Wang, and Cao}]{FANG2024105171_EVA02}
Yuxin Fang, Quan Sun, Xinggang Wang, Tiejun Huang, Xinlong Wang, and Yue Cao. 2024.
\newblock \href {https://doi.org/10.1016/j.imavis.2024.105171} {Eva-02: A visual representation for neon genesis}.
\newblock \emph{Image and Vision Computing}, 149:105171.

\bibitem[{Hamamci et~al.(2024)Hamamci, Er, Almas, Simsek, Esirgun, Dogan, Dasdelen, Wittmann, Simsar, Simsar, Erdemir, Alanbay, Sekuboyina, Lafci, Ozdemir, and Menze}]{hamamci2024ctrate}
Ibrahim~Ethem Hamamci, Sezgin Er, Furkan Almas, Ayse~Gulnihan Simsek, Sevval~Nil Esirgun, Irem Dogan, Muhammed~Furkan Dasdelen, Bastian Wittmann, Enis Simsar, Mehmet Simsar, Emine~Bensu Erdemir, Abdullah Alanbay, Anjany Sekuboyina, Berkan Lafci, Mehmet~K. Ozdemir, and Bjoern Menze. 2024.
\newblock \href {https://arxiv.org/abs/2403.17834} {A foundation model utilizing chest ct volumes and radiology reports for supervised-level zero-shot detection of abnormalities}.
\newblock \emph{Preprint}, arXiv:2403.17834.

\bibitem[{Hu et~al.(2022)Hu, yelong shen, Wallis, Allen-Zhu, Li, Wang, Wang, and Chen}]{hu2022lora}
Edward~J Hu, yelong shen, Phillip Wallis, Zeyuan Allen-Zhu, Yuanzhi Li, Shean Wang, Lu~Wang, and Weizhu Chen. 2022.
\newblock \href {https://openreview.net/forum?id=nZeVKeeFYf9} {Lo{RA}: Low-rank adaptation of large language models}.
\newblock In \emph{International Conference on Learning Representations}.

\bibitem[{Hyland et~al.(2024)Hyland, Bannur, Bouzid, Castro, Ranjit, Schwaighofer, Pérez-García, Salvatelli, Srivastav, Thieme, Codella, Lungren, Wetscherek, Oktay, and Alvarez-Valle}]{hyland2024maira1}
Stephanie~L. Hyland, Shruthi Bannur, Kenza Bouzid, Daniel~C. Castro, Mercy Ranjit, Anton Schwaighofer, Fernando Pérez-García, Valentina Salvatelli, Shaury Srivastav, Anja Thieme, Noel Codella, Matthew~P. Lungren, Maria~Teodora Wetscherek, Ozan Oktay, and Javier Alvarez-Valle. 2024.
\newblock \href {https://arxiv.org/abs/2311.13668} {Maira-1: A specialised large multimodal model for radiology report generation}.
\newblock \emph{Preprint}, arXiv:2311.13668.

\bibitem[{Irvin et~al.(2019)Irvin, Rajpurkar, Ko, Yu, Ciurea-Ilcus, Chute, Marklund, Haghgoo, Ball, Shpanskaya, Seekins, Mong, Halabi, Sandberg, Jones, Larson, Langlotz, Patel, Lungren, and Ng}]{irvin2019chexpert}
Jeremy Irvin, Pranav Rajpurkar, Michael Ko, Yifan Yu, Silviana Ciurea-Ilcus, Chris Chute, Henrik Marklund, Behzad Haghgoo, Robyn Ball, Katie Shpanskaya, Jayne Seekins, David~A. Mong, Safwan~S. Halabi, Jesse~K. Sandberg, Ricky Jones, David~B. Larson, Curtis~P. Langlotz, Bhavik~N. Patel, Matthew~P. Lungren, and Andrew~Y. Ng. 2019.
\newblock \href {https://doi.org/10.1609/aaai.v33i01.3301590} {Chexpert: A large chest radiograph dataset with uncertainty labels and expert comparison}.
\newblock \emph{Proceedings of the AAAI Conference on Artificial Intelligence}, 33(01):590--597.

\bibitem[{Isensee et~al.(2020)Isensee, Jaeger, Kohl, Petersen, and Maier-Hein}]{Isensee_2020_nnunet}
Fabian Isensee, Paul~F. Jaeger, Simon A.~A. Kohl, Jens Petersen, and Klaus~H. Maier-Hein. 2020.
\newblock \href {https://doi.org/10.1038/s41592-020-01008-z} {nnu-net: a self-configuring method for deep learning-based biomedical image segmentation}.
\newblock \emph{Nature Methods}, 18(2):203–211.

\bibitem[{Jain et~al.(2021)Jain, Agrawal, Saporta, Truong, Duong, Bui, Chambon, Zhang, Lungren, Ng, Langlotz, Rajpurkar, and Rajpurkar}]{jain2021radgraph}
Saahil Jain, Ashwin Agrawal, Adriel Saporta, Steven Truong, Du~Nguyen Duong~Nguyen Duong, Tan Bui, Pierre Chambon, Yuhao Zhang, Matthew Lungren, Andrew Ng, Curtis Langlotz, Pranav Rajpurkar, and Pranav Rajpurkar. 2021.
\newblock \href {https://datasets-benchmarks-proceedings.neurips.cc/paper_files/paper/2021/file/c8ffe9a587b126f152ed3d89a146b445-Paper-round1.pdf} {Radgraph: Extracting clinical entities and relations from radiology reports}.
\newblock In \emph{Proceedings of the Neural Information Processing Systems Track on Datasets and Benchmarks}, volume~1.

\bibitem[{Jocobi and Borchardt(1865)}]{Hungarian_Jacobi_1865}
C.G.J. Jocobi and C.W. Borchardt. 1865.
\newblock \href {https://doi.org/doi:10.1515/crll.1865.64.297} {De investigando ordine systematis aequationum differentialium vulgarium cujuscunque.}
\newblock \emph{Journal für die reine und angewandte Mathematik}, 1865(64):297--320.

\bibitem[{Johnson et~al.(2024)Johnson, Lungren, Peng, Lu, Mark, Berkowitz, and Horng}]{johnson2019mimiccxrjpg}
Alistair Johnson, Matt Lungren, Yifan Peng, Zhiyong Lu, Roger Mark, Seth Berkowitz, and Steven Horng. 2024.
\newblock Mimic-cxr-jpg - chest radiographs with structured labels (version 2.1.0).
\newblock \emph{PhysioNet}.

\bibitem[{Johnson et~al.(2019)Johnson, Pollard, Berkowitz, Greenbaum, Lungren, Deng, Mark, and Horng}]{johnson2019mimiccxr}
Alistair~EW Johnson, Tom~J Pollard, Seth~J Berkowitz, Nathaniel~R Greenbaum, Matthew~P Lungren, Chih-ying Deng, Roger~G Mark, and Steven Horng. 2019.
\newblock Mimic-cxr, a de-identified publicly available database of chest radiographs with free-text reports.
\newblock \emph{Scientific data}, 6(1):317.

\bibitem[{Kalajdzievski(2023)}]{kalajdzievski2023rankstabilizationscalingfactor}
Damjan Kalajdzievski. 2023.
\newblock \href {https://arxiv.org/abs/2312.03732} {A rank stabilization scaling factor for fine-tuning with lora}.
\newblock \emph{Preprint}, arXiv:2312.03732.

\bibitem[{Kirillov et~al.(2023)Kirillov, Mintun, Ravi, Mao, Rolland, Gustafson, Xiao, Whitehead, Berg, Lo, Dollár, and Girshick}]{sam2023iccv}
Alexander Kirillov, Eric Mintun, Nikhila Ravi, Hanzi Mao, Chloe Rolland, Laura Gustafson, Tete Xiao, Spencer Whitehead, Alexander~C. Berg, Wan-Yen Lo, Piotr Dollár, and Ross Girshick. 2023.
\newblock \href {https://doi.org/10.1109/ICCV51070.2023.00371} {Segment anything}.
\newblock In \emph{2023 IEEE/CVF International Conference on Computer Vision (ICCV)}, pages 3992--4003.

\bibitem[{Kuhn(1955)}]{Kuhn1955hungarian}
H.~W. Kuhn. 1955.
\newblock \href {https://doi.org/10.1002/nav.3800020109} {The hungarian method for the assignment problem}.
\newblock \emph{Naval Research Logistics Quarterly}, 2(1-2):83--97.

\bibitem[{Lai et~al.(2023)Lai, Tian, Chen, Li, Yuan, Liu, and Jia}]{lai2023lisa}
Xin Lai, Zhuotao Tian, Yukang Chen, Yanwei Li, Yuhui Yuan, Shu Liu, and Jiaya Jia. 2023.
\newblock Lisa: Reasoning segmentation via large language model.
\newblock \emph{arXiv preprint arXiv:2308.00692}.

\bibitem[{Lau et~al.(2018)Lau, Gayen, Ben~Abacha, and Demner-Fushman}]{lau2018vqarad}
Jason~J Lau, Soumya Gayen, Asma Ben~Abacha, and Dina Demner-Fushman. 2018.
\newblock A dataset of clinically generated visual questions and answers about radiology images.
\newblock \emph{Scientific data}, 5(1):1--10.

\bibitem[{Lefaudeux et~al.(2022)Lefaudeux, Massa, Liskovich, Xiong, Caggiano, Naren, Xu, Hu, Tintore, Zhang, Labatut, Haziza, Wehrstedt, Reizenstein, and Sizov}]{xFormers2022}
Benjamin Lefaudeux, Francisco Massa, Diana Liskovich, Wenhan Xiong, Vittorio Caggiano, Sean Naren, Min Xu, Jieru Hu, Marta Tintore, Susan Zhang, Patrick Labatut, Daniel Haziza, Luca Wehrstedt, Jeremy Reizenstein, and Grigory Sizov. 2022.
\newblock xformers: A modular and hackable transformer modelling library.
\newblock \url{https://github.com/facebookresearch/xformers}.

\bibitem[{Li et~al.(2023)Li, Wong, Zhang, Usuyama, Liu, Yang, Naumann, Poon, and Gao}]{li2023llavamed}
Chunyuan Li, Cliff Wong, Sheng Zhang, Naoto Usuyama, Haotian Liu, Jianwei Yang, Tristan Naumann, Hoifung Poon, and Jianfeng Gao. 2023.
\newblock Llava-med: Training a large language-and-vision assistant for biomedicine in one day.
\newblock \emph{arXiv preprint arXiv:2306.00890}.

\bibitem[{Lin(2004)}]{lin2004rouge}
Chin-Yew Lin. 2004.
\newblock Rouge: A package for automatic evaluation of summaries.
\newblock In \emph{Text summarization branches out}, pages 74--81.

\bibitem[{Lin et~al.(2020)Lin, Goyal, Girshick, He, and Dollár}]{focal_loss_TPAMI_2020_Lin}
Tsung-Yi Lin, Priya Goyal, Ross Girshick, Kaiming He, and Piotr Dollár. 2020.
\newblock \href {https://doi.org/10.1109/TPAMI.2018.2858826} {Focal loss for dense object detection}.
\newblock \emph{IEEE Transactions on Pattern Analysis and Machine Intelligence}, 42(2):318--327.

\bibitem[{Liu et~al.(2021)Liu, Zhan, Xu, Ma, Yang, and Wu}]{liu2021slake}
Bo~Liu, Li-Ming Zhan, Li~Xu, Lin Ma, Yan Yang, and Xiao-Ming Wu. 2021.
\newblock Slake: A semantically-labeled knowledge-enhanced dataset for medical visual question answering.
\newblock In \emph{2021 IEEE 18th International Symposium on Biomedical Imaging (ISBI)}, pages 1650--1654. IEEE.

\bibitem[{Liu et~al.(2024)Liu, Li, Li, Li, Zhang, Shen, and Lee}]{liu2024llavanext}
Haotian Liu, Chunyuan Li, Yuheng Li, Bo~Li, Yuanhan Zhang, Sheng Shen, and Yong~Jae Lee. 2024.
\newblock \href {https://llava-vl.github.io/blog/2024-01-30-llava-next/} {Llava-next: Improved reasoning, ocr, and world knowledge}.

\bibitem[{Luo et~al.(2024)Luo, Chen, Tang, Chen, Han, Hu, Li, and Chen}]{SpadNets_arxiv_2024_Luo}
Lingxiao Luo, Xuanzhong Chen, Bingda Tang, Xinsheng Chen, Rong Han, Chengpeng Hu, Yujiang Li, and Ting Chen. 2024.
\newblock \href {https://arxiv.org/abs/2312.07630} {Building universal foundation models for medical image analysis with spatially adaptive networks}.
\newblock \emph{Preprint}, arXiv:2312.07630.

\bibitem[{Moor et~al.(2023)Moor, Banerjee, Abad, Krumholz, Leskovec, Topol, and Rajpurkar}]{Moor2023}
Michael Moor, Oishi Banerjee, Zahra Shakeri~Hossein Abad, Harlan~M. Krumholz, Jure Leskovec, Eric~J. Topol, and Pranav Rajpurkar. 2023.
\newblock \href {https://doi.org/10.1038/s41586-023-05881-4} {Foundation models for generalist medical artificial intelligence}.
\newblock \emph{Nature}, 616(7956):259--265.

\bibitem[{Papineni et~al.(2002)Papineni, Roukos, Ward, and Zhu}]{papineni2002bleu}
Kishore Papineni, Salim Roukos, Todd Ward, and Wei-Jing Zhu. 2002.
\newblock Bleu: a method for automatic evaluation of machine translation.
\newblock In \emph{Proceedings of the 40th annual meeting of the Association for Computational Linguistics}, pages 311--318.

\bibitem[{Peng et~al.(2024)Peng, Wang, Dong, Hao, Huang, Ma, Ye, and Wei}]{peng2024grounding}
Zhiliang Peng, Wenhui Wang, Li~Dong, Yaru Hao, Shaohan Huang, Shuming Ma, Qixiang Ye, and Furu Wei. 2024.
\newblock \href {https://openreview.net/forum?id=lLmqxkfSIw} {Grounding multimodal large language models to the world}.
\newblock In \emph{The Twelfth International Conference on Learning Representations}.

\bibitem[{Pi et~al.(2023)Pi, Gao, Diao, Pan, Dong, Zhang, Yao, Han, Xu, Kong, and Zhang}]{pi-etal-2023-detgpt}
Renjie Pi, Jiahui Gao, Shizhe Diao, Rui Pan, Hanze Dong, Jipeng Zhang, Lewei Yao, Jianhua Han, Hang Xu, Lingpeng Kong, and Tong Zhang. 2023.
\newblock \href {https://doi.org/10.18653/v1/2023.emnlp-main.876} {{D}et{GPT}: Detect what you need via reasoning}.
\newblock In \emph{Proceedings of the 2023 Conference on Empirical Methods in Natural Language Processing}, pages 14172--14189, Singapore. Association for Computational Linguistics.

\bibitem[{Rasheed et~al.(2024)Rasheed, Maaz, Shaji, Shaker, Khan, Cholakkal, Anwer, Xing, Yang, and Khan}]{Rasheed_2024_CVPR_GLaMM}
Hanoona Rasheed, Muhammad Maaz, Sahal Shaji, Abdelrahman Shaker, Salman Khan, Hisham Cholakkal, Rao~M. Anwer, Eric Xing, Ming-Hsuan Yang, and Fahad~S. Khan. 2024.
\newblock Glamm: Pixel grounding large multimodal model.
\newblock In \emph{Proceedings of the IEEE/CVF Conference on Computer Vision and Pattern Recognition (CVPR)}, pages 13009--13018.

\bibitem[{Ren et~al.(2023)Ren, Liu, Li, Zhang, Zeng, Yang, Liao, Jia, Li, Cao, Wang, Zeng, Qi, Yuan, Yang, and Zhang}]{detrex}
Tianhe Ren, Shilong Liu, Feng Li, Hao Zhang, Ailing Zeng, Jie Yang, Xingyu Liao, Ding Jia, Hongyang Li, He~Cao, Jianan Wang, Zhaoyang Zeng, Xianbiao Qi, Yuhui Yuan, Jianwei Yang, and Lei Zhang. 2023.
\newblock \href {https://doi.org/10.48550/ARXIV.2306.07265} {detrex: Benchmarking detection transformers}.
\newblock \emph{arXiv preprint}.

\bibitem[{Rezatofighi et~al.(2019)Rezatofighi, Tsoi, Gwak, Sadeghian, Reid, and Savarese}]{GIoU_CVPR_2019_Rezatofighi}
Hamid Rezatofighi, Nathan Tsoi, JunYoung Gwak, Amir Sadeghian, Ian Reid, and Silvio Savarese. 2019.
\newblock \href {https://doi.org/10.1109/CVPR.2019.00075} {Generalized intersection over union: A metric and a loss for bounding box regression}.
\newblock In \emph{2019 IEEE/CVF Conference on Computer Vision and Pattern Recognition (CVPR)}, pages 658--666.

\bibitem[{Rückert et~al.(2024)Rückert, Bloch, Brüngel, Idrissi-Yaghir, Schäfer, Schmidt, Koitka, Pelka, Abacha, de~Herrera, Müller, Horn, Nensa, and Friedrich}]{rückert2024rocov2}
Johannes Rückert, Louise Bloch, Raphael Brüngel, Ahmad Idrissi-Yaghir, Henning Schäfer, Cynthia~S. Schmidt, Sven Koitka, Obioma Pelka, Asma~Ben Abacha, Alba G.~Seco de~Herrera, Henning Müller, Peter~A. Horn, Felix Nensa, and Christoph~M. Friedrich. 2024.
\newblock \href {https://arxiv.org/abs/2405.10004} {Rocov2: Radiology objects in context version 2, an updated multimodal image dataset}.
\newblock \emph{Preprint}, arXiv:2405.10004.

\bibitem[{Shazeer(2020)}]{shazeer2020gluvariantsimprovetransformer}
Noam Shazeer. 2020.
\newblock \href {https://arxiv.org/abs/2002.05202} {Glu variants improve transformer}.
\newblock \emph{Preprint}, arXiv:2002.05202.

\bibitem[{Smit et~al.(2020)Smit, Jain, Rajpurkar, Pareek, Ng, and Lungren}]{smit2020chexbert}
Akshay Smit, Saahil Jain, Pranav Rajpurkar, Anuj Pareek, Andrew Ng, and Matthew Lungren. 2020.
\newblock \href {https://doi.org/10.18653/v1/2020.emnlp-main.117} {Combining automatic labelers and expert annotations for accurate radiology report labeling using {BERT}}.
\newblock In \emph{Proceedings of the 2020 Conference on Empirical Methods in Natural Language Processing (EMNLP)}, pages 1500--1519, Online. Association for Computational Linguistics.

\bibitem[{Wang et~al.(2023{\natexlab{a}})Wang, Lv, Yu, Hong, Qi, Wang, Ji, Yang, Zhao, Song et~al.}]{wang2023cogvlm}
Weihan Wang, Qingsong Lv, Wenmeng Yu, Wenyi Hong, Ji~Qi, Yan Wang, Junhui Ji, Zhuoyi Yang, Lei Zhao, Xixuan Song, et~al. 2023{\natexlab{a}}.
\newblock Cogvlm: Visual expert for pretrained language models.
\newblock \emph{arXiv preprint arXiv:2311.03079}.

\bibitem[{Wang et~al.(2023{\natexlab{b}})Wang, Liu, Wang, and Zhou}]{wang2023r2gengpt}
Zhanyu Wang, Lingqiao Liu, Lei Wang, and Luping Zhou. 2023{\natexlab{b}}.
\newblock \href {https://doi.org/10.1016/j.metrad.2023.100033} {R2gengpt: Radiology report generation with frozen llms}.
\newblock \emph{Meta-Radiology}, 1(3):100033.

\bibitem[{Wasserthal et~al.(2023)Wasserthal, Breit, Meyer, Pradella, Hinck, Sauter, Heye, Boll, Cyriac, Yang, Bach, and Segeroth}]{totalsegmentator}
Jakob Wasserthal, Hanns-Christian Breit, Manfred~T. Meyer, Maurice Pradella, Daniel Hinck, Alexander~W. Sauter, Tobias Heye, Daniel~T. Boll, Joshy Cyriac, Shan Yang, Michael Bach, and Martin Segeroth. 2023.
\newblock \href {https://doi.org/10.1148/ryai.230024} {Totalsegmentator: Robust segmentation of 104 anatomic structures in ct images}.
\newblock \emph{Radiology: Artificial Intelligence}, 5(5):e230024.

\bibitem[{Wu et~al.(2023{\natexlab{a}})Wu, Lei, Zheng, Zhao, Lin, Zhang, Zhou, Zhao, Zhang, Wang et~al.}]{wu2023can}
Chaoyi Wu, Jiayu Lei, Qiaoyu Zheng, Weike Zhao, Weixiong Lin, Xiaoman Zhang, Xiao Zhou, Ziheng Zhao, Ya~Zhang, Yanfeng Wang, et~al. 2023{\natexlab{a}}.
\newblock Can gpt-4v (ision) serve medical applications? case studies on gpt-4v for multimodal medical diagnosis.
\newblock \emph{arXiv preprint arXiv:2310.09909}.

\bibitem[{Wu et~al.(2023{\natexlab{b}})Wu, Zhang, Zhang, Wang, and Xie}]{wu2023radfm}
Chaoyi Wu, Xiaoman Zhang, Ya~Zhang, Yanfeng Wang, and Weidi Xie. 2023{\natexlab{b}}.
\newblock \href {https://arxiv.org/abs/2308.02463} {Towards generalist foundation model for radiology by leveraging web-scale 2d\&3d medical data}.
\newblock \emph{Preprint}, arXiv:2308.02463.

\bibitem[{Yang et~al.(2024)Yang, Xu, Sellergren, Kohlberger, Zhou, Ktena, Kiraly, Ahmed, Hormozdiari, Jaroensri, Wang, Wulczyn, Jamil, Guidroz, Lau, Qiao, Liu, Goel, Park, Agharwal, George, Wang, Tanno, Barrett, Weng, Mahdavi, Saab, Tu, Kalidindi, Etemadi, Cuadros, Sorensen, Matias, Chou, Corrado, Barral, Shetty, Fleet, Eslami, Tse, Prabhakara, McLean, Steiner, Pilgrim, Kelly, Azizi, and Golden}]{yang2024medgemini}
Lin Yang, Shawn Xu, Andrew Sellergren, Timo Kohlberger, Yuchen Zhou, Ira Ktena, Atilla Kiraly, Faruk Ahmed, Farhad Hormozdiari, Tiam Jaroensri, Eric Wang, Ellery Wulczyn, Fayaz Jamil, Theo Guidroz, Chuck Lau, Siyuan Qiao, Yun Liu, Akshay Goel, Kendall Park, Arnav Agharwal, Nick George, Yang Wang, Ryutaro Tanno, David G.~T. Barrett, Wei-Hung Weng, S.~Sara Mahdavi, Khaled Saab, Tao Tu, Sreenivasa~Raju Kalidindi, Mozziyar Etemadi, Jorge Cuadros, Gregory Sorensen, Yossi Matias, Katherine Chou, Greg Corrado, Joelle Barral, Shravya Shetty, David Fleet, S.~M.~Ali Eslami, Daniel Tse, Shruthi Prabhakara, Cory McLean, Dave Steiner, Rory Pilgrim, Christopher Kelly, Shekoofeh Azizi, and Daniel Golden. 2024.
\newblock \href {https://arxiv.org/abs/2405.03162} {Advancing multimodal medical capabilities of gemini}.
\newblock \emph{Preprint}, arXiv:2405.03162.

\bibitem[{Yang et~al.(2023)Yang, Qu, Lai, Tian, Peng, Liu, and Jia}]{yang2023improved}
Senqiao Yang, Tianyuan Qu, Xin Lai, Zhuotao Tian, Bohao Peng, Shu Liu, and Jiaya Jia. 2023.
\newblock An improved baseline for reasoning segmentation with large language model.
\newblock \emph{arXiv preprint arXiv:2312.17240}.

\bibitem[{You et~al.(2024)You, Zhang, Gan, Du, Zhang, Wang, Cao, Chang, and Yang}]{you2024ferret}
Haoxuan You, Haotian Zhang, Zhe Gan, Xianzhi Du, Bowen Zhang, Zirui Wang, Liangliang Cao, Shih-Fu Chang, and Yinfei Yang. 2024.
\newblock \href {https://openreview.net/forum?id=2msbbX3ydD} {Ferret: Refer and ground anything anywhere at any granularity}.
\newblock In \emph{The Twelfth International Conference on Learning Representations}.

\bibitem[{Yu et~al.(2023)Yu, Endo, Krishnan, Pan, Tsai, Reis, Fonseca, Lee, Abad, Ng et~al.}]{yu2023radcliq}
Feiyang Yu, Mark Endo, Rayan Krishnan, Ian Pan, Andy Tsai, Eduardo~Pontes Reis, Eduardo Kaiser Ururahy~Nunes Fonseca, Henrique Min~Ho Lee, Zahra Shakeri~Hossein Abad, Andrew~Y Ng, et~al. 2023.
\newblock Evaluating progress in automatic chest x-ray radiology report generation.
\newblock \emph{Patterns}, 4(9).

\bibitem[{Zhang et~al.(2023)Zhang, Li, Liu, Zhang, Su, Zhu, Ni, and Shum}]{DINO_ICLR_2023_Zhang}
Hao Zhang, Feng Li, Shilong Liu, Lei Zhang, Hang Su, Jun Zhu, Lionel Ni, and Heung-Yeung Shum. 2023.
\newblock \href {https://openreview.net/forum?id=3mRwyG5one} {{DINO}: {DETR} with improved denoising anchor boxes for end-to-end object detection}.
\newblock In \emph{The Eleventh International Conference on Learning Representations}.

\bibitem[{Zhang et~al.(2019)Zhang, Kishore, Wu, Weinberger, and Artzi}]{zhang2019bertscore}
Tianyi Zhang, Varsha Kishore, Felix Wu, Kilian~Q Weinberger, and Yoav Artzi. 2019.
\newblock Bertscore: Evaluating text generation with bert.
\newblock In \emph{International Conference on Learning Representations}.

\bibitem[{Zhang et~al.(2024)Zhang, Ma, Gao, Shakiah, Gao, and Chai}]{zhang2024groundhog}
Yichi Zhang, Ziqiao Ma, Xiaofeng Gao, Suhaila Shakiah, Qiaozi Gao, and Joyce Chai. 2024.
\newblock Groundhog: Grounding large language models to holistic segmentation.
\newblock \emph{arXiv preprint arXiv:2402.16846}.

\bibitem[{Zhao et~al.(2024)Zhao, Zhang, Wu, Zhang, Zhang, Wang, and Xie}]{zhao2024modelrulealluniversal}
Ziheng Zhao, Yao Zhang, Chaoyi Wu, Xiaoman Zhang, Ya~Zhang, Yanfeng Wang, and Weidi Xie. 2024.
\newblock \href {https://arxiv.org/abs/2312.17183} {One model to rule them all: Towards universal segmentation for medical images with text prompts}.
\newblock \emph{Preprint}, arXiv:2312.17183.

\bibitem[{Zhou et~al.(2024)Zhou, Ong, Kennedy, Wu, Kazam, Hentel, Flanders, Shih, and Peng}]{zhou2024evaluating}
Yiliang Zhou, Hanley Ong, Patrick Kennedy, Carol~C Wu, Jacob Kazam, Keith Hentel, Adam Flanders, George Shih, and Yifan Peng. 2024.
\newblock Evaluating gpt-v4 (gpt-4 with vision) on detection of radiologic findings on chest radiographs.
\newblock \emph{Radiology}, 311(2):e233270.

\end{thebibliography}

\appendix

\section{Symbols}

The description of symbols used in this manuscript are listed in Table~\ref{tab:symbols}.

\begin{table*}[h!]
\centering
\begin{tabularx}{\linewidth}{lX}
\toprule
\textbf{Symbol} & \textbf{Description} \\
\midrule
\(S_m\) & The set of all permutations with \(m\) elements, or commonly known as the symmetric group of order \(m\)\\
\(\sigma\) & A permutation (an element of \(S_m\) for some \(m\)), i.e., a bijection that maps from a finite set with \(m\) elements to itself \\
\(y_i\) & The \(i\)-th ground truth bounding box label, where \(y\) may be padded with dummy negative instances \\
\(\hat{y}_i\) & The \(i\)-th bounding box predicted by the model \\
\(L_{\mathrm{cost}}(\cdot, \cdot)\) & The cost function matching a ground truth with a prediction, used by the Hungarian algorithm for weighted bipartite matching \\
\(c_i\) & Indicates if \(y_i\) is a dummy negative instance (\(c_i=0\)) or not (\(c_i=1\)) \\
\(\hat{p}_i\) & The predicted probability by the model that \(\hat{y}_i\) is positive (not a padded dummy negative instance) \\
\(L_{\mathrm{box}}(\cdot, \cdot)\) & The loss function for bounding box regression; a component of \(L_{\mathrm{cost}}\) \\
\(L_{\mathrm{disc}}(\cdot, \cdot)\) & The loss function for positive/negative classification for predicted instances; a component of \(L_{\mathrm{cost}}\) \\
\(a \uparrow b\) & Knuth's up-arrow notation for exponentiation, equivalent to \(a^b\) \\
\bottomrule
\end{tabularx}
\caption{Description of symbols.}
\label{tab:symbols}
\end{table*}

\section{Base VLM Architecture}
\label{sec:vlm}

In brief, CogVLM 17B consists of a vision transformer (ViT) encoder, an MLP adapter of the SwiGLU variant~\cite{shazeer2020gluvariantsimprovetransformer}, an LLM based on Vicuna 1.5 7B ~\cite{vicuna2023}. 

During inference, firstly the ViT encoder divides the input image into non-overlapping patches and encodes them as image embeddings.
Then the MLP adapter is employed to project the image embeddings into the language embedding space.
Finally, the LLM generates responses by processing the concatenated projected image embeddings and the embeddings of language instructions.
Notably, the image embeddings are processed using a separate set of parameters within the transformer layers of the LLM, originally referred to as the visual expert module. The visual expert module parameters are initialized from the pretrained LLM.

\section{Clinical Metrics for Reports}
\label{sec:clinical-report-metrics}

\subsection{MIMIC-CXR}

For the chest X-ray images from the MIMIC-CXR dataset, the following metrics are computed.
\paragraph{CheXpert F1 and FNR}

The macro and micro F1 scores and false nagative rates (FNRs) averaged over all 14 and 5 major\footnote{Following previous works\cite{hyland2024maira1}, the 5 major categories considered are: atelectasis, cardiomegaly, consolidation, edema, and pleural effusion.} CheXpert pathological observations~\cite{irvin2019chexpert} extracted from the generated and reference reports using the CheXbert model~\cite{smit2020chexbert}.

\paragraph{CheXbert vector similarity}

The cosine similarity between the CheXbert-embedded reference and generated reports.

\paragraph{RadGraph F1}

The F1 score for the presence of clinical entities and their relations extracted by the RadGraph model~\cite{jain2021radgraph}.

\paragraph{RadCliQ v1}~\cite{yu2023radcliq}: A composite metric integrating BLEU-2, CheXbert vector similarity, RadGraph F1, and BERTScore~\cite{zhang2019bertscore} to predict report errors.

\subsection{CT-RATE}

For the CT-RATE dataset, we compute RadBERT F1 and FNR, which is analogous to CheXpert F1 and FNR, but uses the RadBERT model trained specifically on the CT-RATE dataset to extract 18 abnormalities, which is used to annotate CT-RATE~\cite{hamamci2024ctrate}.

\section{Discussion on Diverse Input Handling}
\label{sec:discuss-input-handling}

Conventional vision models require all input tensors have the same spatial dimensions.
Interpolation is a common technique to fulfill purpose.
However, 3D medical images can have varying sizes, and inter-slice interpolation can introduce significant artifacts, particularly in regions where anatomical continuity is crucial.
When interpolating these images to a unified size or voxel spacing, there is a risk of losing important spatial information. As a result, inter-slice interpolation can blur critical details and introduce uncertainty in tasks such as segmentation and detection, ultimately affecting the clinical interpretation of medical images.

In contrast, our proposed approach leverages dynamic patch embeddings, which adapt to the specific characteristics of each image without requiring uniform interpolation. 
This allows us to maintain the original spatial resolution and anatomical integrity of the image, ensuring that fine details are preserved.
The principle of avoiding interpolation but adapting the (static) model architecture to image properties was empirically validated by nnU-Net by segmentation performance, and is widely used by nowadays medical image analysis models.

\section{Hallucination Vulnerability Addressing}
\label{sec:data-cleaning}

Radiologists frequently reference external information in reports, such as images of other views, prior examinations, and the patient's medical history.
While being crucial for diagnosis, such information cannot be inferred solely from a single image, and models trained on such reports tend to hallucinate, generating unfounded references to nonexistent external information~\cite{hyland2024maira1}.
While this issue should be resolved by including enough information~\cite{bannur2024maira2groundedradiologyreport},
such as images from other views, images and reports from prior studies images, prior reports, and the ``Indication'', ``Technique'' and ``Comparison'' sections of the current report into the input~\cite{bannur2024maira2groundedradiologyreport}.
However,
such approach is infeasible for most datasets as they still fail to cover all external information of concern, especially for existing open datasets such as MIMIC-CXR and CT-RATE.

Conversely, we opt for a more flexible approach that removes content in the report that may result in hallucinations.
Specifically, we instruct Meta Llama 3 70B~\cite{dubey2024llama3herdmodels} to process reports and captions by removing all references to external information.
Meanwhile, sentences are paraphrased based on the context with minimal modifications to minimize distribution shift and information loss.

\section{Implementation Details}

The details of hyperparameter settings are presented in Table~\ref{table:hp-pt}.
We train \modelname{} for 40k, 50k, and 10k steps for 3 stages, respectively.
We adopt rank-stabilized LoRA~(rsLoRA)~\cite{hu2022lora, kalajdzievski2023rankstabilizationscalingfactor} with rank = 64 and \(\alpha\)~= 8 to adapt from the pre-trained general-purpose VLM.
The base patch size of ViT is 16 for all spatial dimensions.
Inspired by M3D~\cite{bai2024m3d}, we adopt a max pooling layer to reduce the spatial dimensions of feature maps output by the vision encoder by a factor of 2 when applicable.

We implement our model largely based on PyTorch 2~\cite{pytorch2} and MONAI~\cite{cardoso2022monaiopensourceframeworkdeep}.
Additionally, we adapt the vision encoder and LLM to using FlashAttention-2~\cite{flashattn, dao2024flashattention2} for computation efficiency based on the xFormers~\cite{xFormers2022}.
The DINO model for disease detection on chest X-ray images is trained based on the detrex library~\cite{detrex}.
Our models are trained on 8 NVIDIA A100 GPUs with 80 GB memory.

For all downstream tasks, we train all models for the same batch size, number of iterations, and learning rate schedule. 
We adapt the open-sourced implementation of baselines to our settings.
We refer readers to the detailed configuration in our code for each task and model.

\begin{table*}[htbp]
    \centering
    \begin{tabular}{l|ccc}
    \toprule
    Configuration       & Stage 1             & Stage 2            & Stage 3           \\
    \midrule
    training steps      & 40k                 & 50k                & 10k               \\
    linear warmup steps & 2k                  & 2.5k               & 0                 \\
    batch size          & \multicolumn{3}{c}{128}                                      \\
    peak lr             & 5e-5                & 5e-5               & 2e-5              \\
    lr schedule         & \multicolumn{3}{c}{cosine decay}                             \\
    grad. clip norm.    & \multicolumn{3}{c}{1}                                        \\
    optimizer           & \multicolumn{3}{c}{AdamW}                                    \\
    Adam parameters     & \multicolumn{3}{c}{\(\beta = (0.9, 0.999), \varepsilon = \text{1e-8}\)} \\
    weight decay        & \multicolumn{3}{c}{5e-2}                                     \\
    base LLM            & \multicolumn{3}{c}{Vicuna-1.5-7B}                            \\
    base ViT patch size & \multicolumn{3}{c}{\(16 \times 16 \times 16\)}                                 \\
    numerical precision & \multicolumn{3}{c}{bfloat16}                                 \\
    LoRA rank           & \multicolumn{3}{c}{64}                                       \\
    LoRA \(\alpha\)     & \multicolumn{3}{c}{8}                                        \\
    LoRA dropout        & \multicolumn{3}{c}{0.05}                                     \\
    rsLoRA           & \multicolumn{3}{c}{\checkmark}                                \\
    \bottomrule
    \end{tabular}
    \caption{Hyperparameter settings of \modelname{}.}
    \label{table:hp-pt}
\end{table*}

\section{More Qualitative Examples}
\label{sec:grg-qualitative}

\begin{figure*}[htbp]
    \centering
    \begin{subfigure}[b]{0.45\linewidth}
        \centering
        \includegraphics[width=\linewidth]{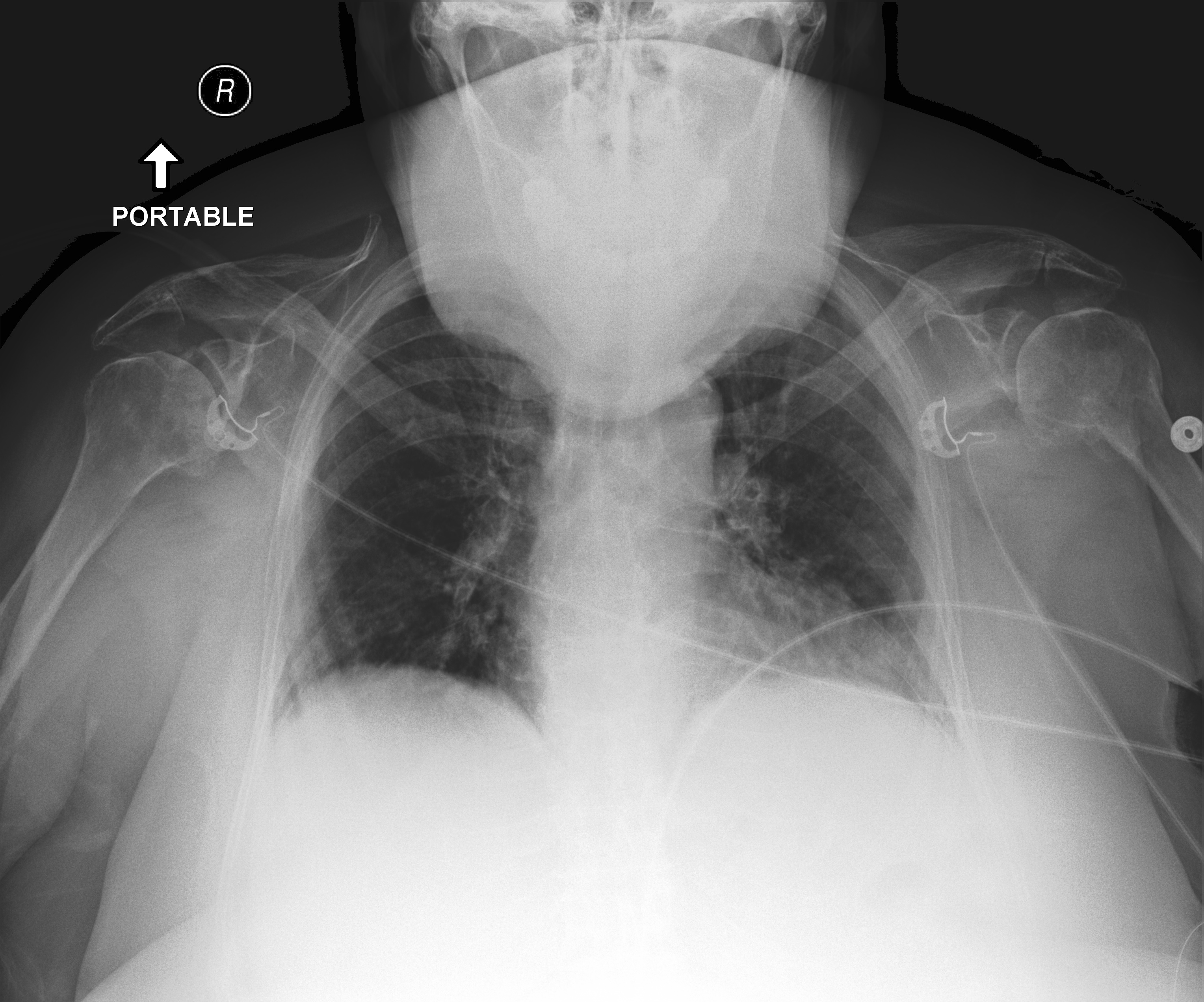}
    \end{subfigure}
    \hspace{0.03\linewidth}
    \begin{subfigure}[b]{0.45\linewidth}
        \centering
        \includegraphics[width=\linewidth]{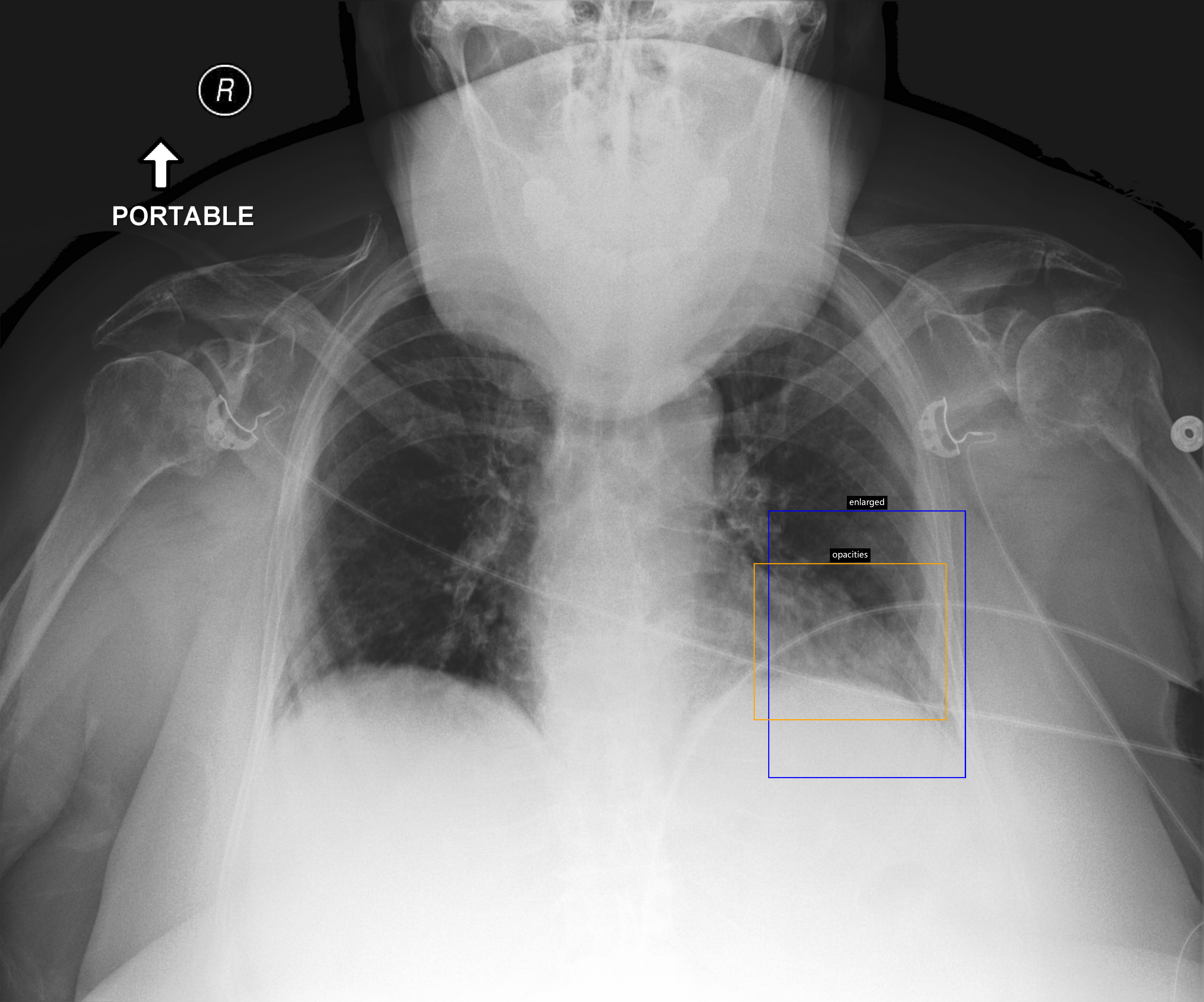}
    \end{subfigure}
    \begin{tcolorbox}
    \textbf{\modelname{}}:
    Findings: Lung volumes are low.  Heart size is mildly {\color{blue} enlarged}. Mediastinal and hilar contours are unremarkable.  Pulmonary vasculature is not engorged.  Patchy {\color{orange} opacities} are noted in the lung bases.  No pleural effusion or pneumothorax is present.  No acute osseous abnormalities detected.
    
    Impression: Low lung volumes with patchy bibasilar airspace opacities, likely atelectasis in the setting of chronic obstructive pulmonary disease.
    \end{tcolorbox}
    \begin{tcolorbox}
    \textbf{Reference}:
    Findings: The lung volumes are low.  A patchy left basilar opacity obscures the cardiac border and apex of the left hemidiaphragm, worrisome for pneumonia.  Elsewhere, the lungs appear clear.  There are no pleural effusions or pneumothorax.

    Impression: Left basilar opacity worrisome for pneumonia.
    \end{tcolorbox}
    \caption{
        In this example, the model wrongly identifies cardiomegaly and gives an unusual visual grounding result, which may remind the radiologist in clinical practice.
    }
    \label{fig:more-grg-1}
\end{figure*}

\begin{figure*}[htbp]
    \centering
    \begin{subfigure}[b]{0.45\linewidth}
        \centering
        \includegraphics[width=\linewidth]{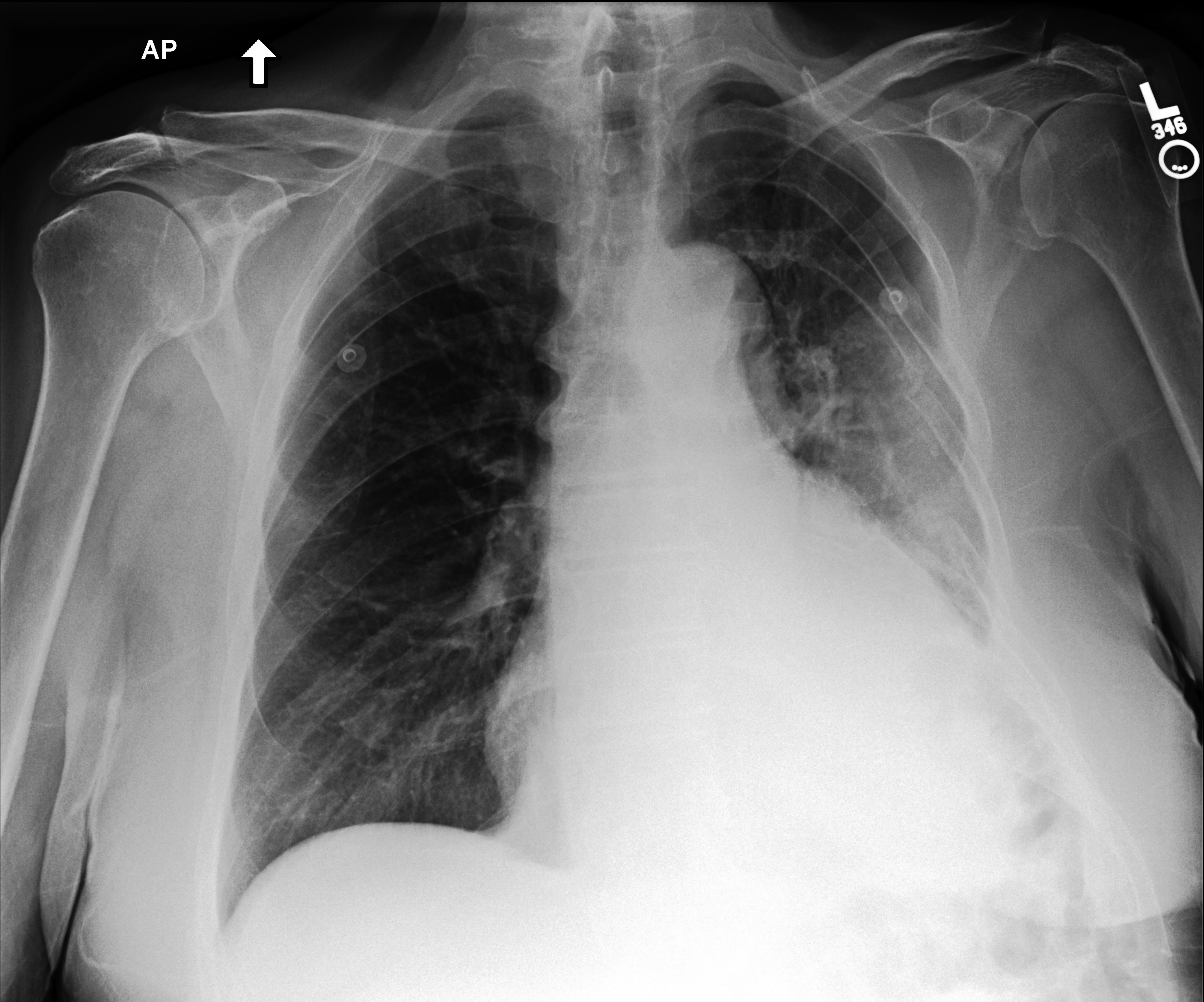}
    \end{subfigure}
    \hspace{0.03\linewidth}
    \begin{subfigure}[b]{0.45\linewidth}
        \centering
        \includegraphics[width=\linewidth]{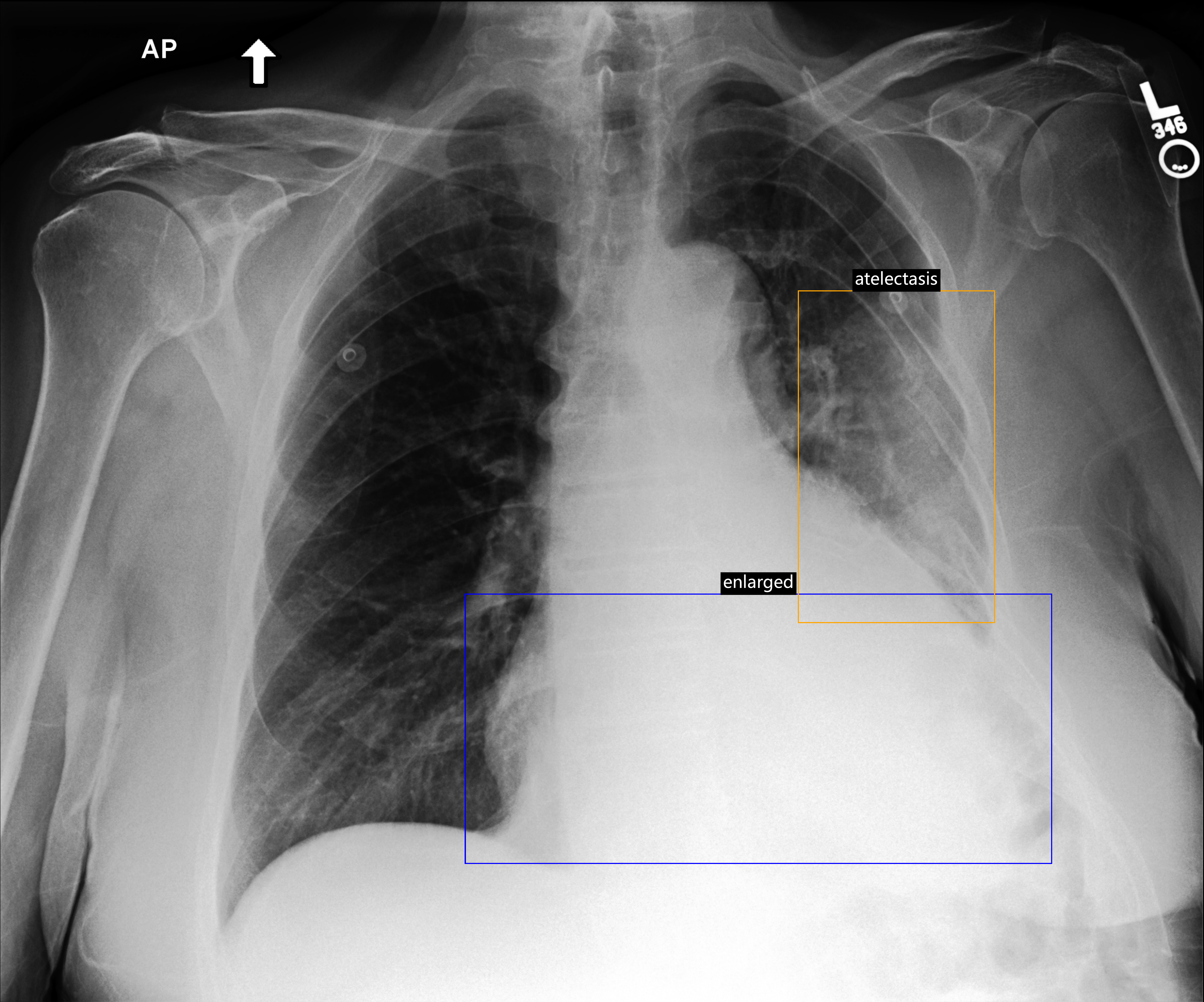}
    \end{subfigure}
    \begin{tcolorbox}
    \textbf{\modelname{}}:
    Findings: Heart size is \textcolor{blue}{enlarged}. Left lower lobe \textcolor{orange}{atelectasis} is present. Right lung is clear. Mediastinal contour is normal.  Bony structures are intact. No evidence for pleural effusion.

    Impression: Cardiomegaly with left lower lobe atelectasis.
    \end{tcolorbox}
    \begin{tcolorbox}
    \textbf{Reference}:
    Findings: The heart size is mildly enlarged.  The aorta is tortuous.  Elevation of the left hemidiaphragm is seen.  A left mid posterior chest wall deformity is demonstrated  with associated right basilar \textcolor{red}{opacity}.  There is left basilar atelectasis.  The right lung is clear.  No pleural effusion or pneumothorax is seen.  There is no pulmonary vascular congestion.  Mild degenerative changes are noted in the thoracic spine.

    Impression: Postoperative appearance of the left chest without acute cardiopulmonary abnormality.
    \end{tcolorbox}
    \caption{
        In this example, the model correctly identifies cardiomegaly and atelectasis, validated by corresponding bounding boxes output by the visual grounding.
        However, it omits the presented \textcolor{red}{opacity}.
    }
    \label{fig:more-grg-2}
\end{figure*}

\begin{figure*}[htbp]
    \centering
    \begin{subfigure}[b]{0.45\linewidth}
        \centering
        \includegraphics[width=\linewidth]{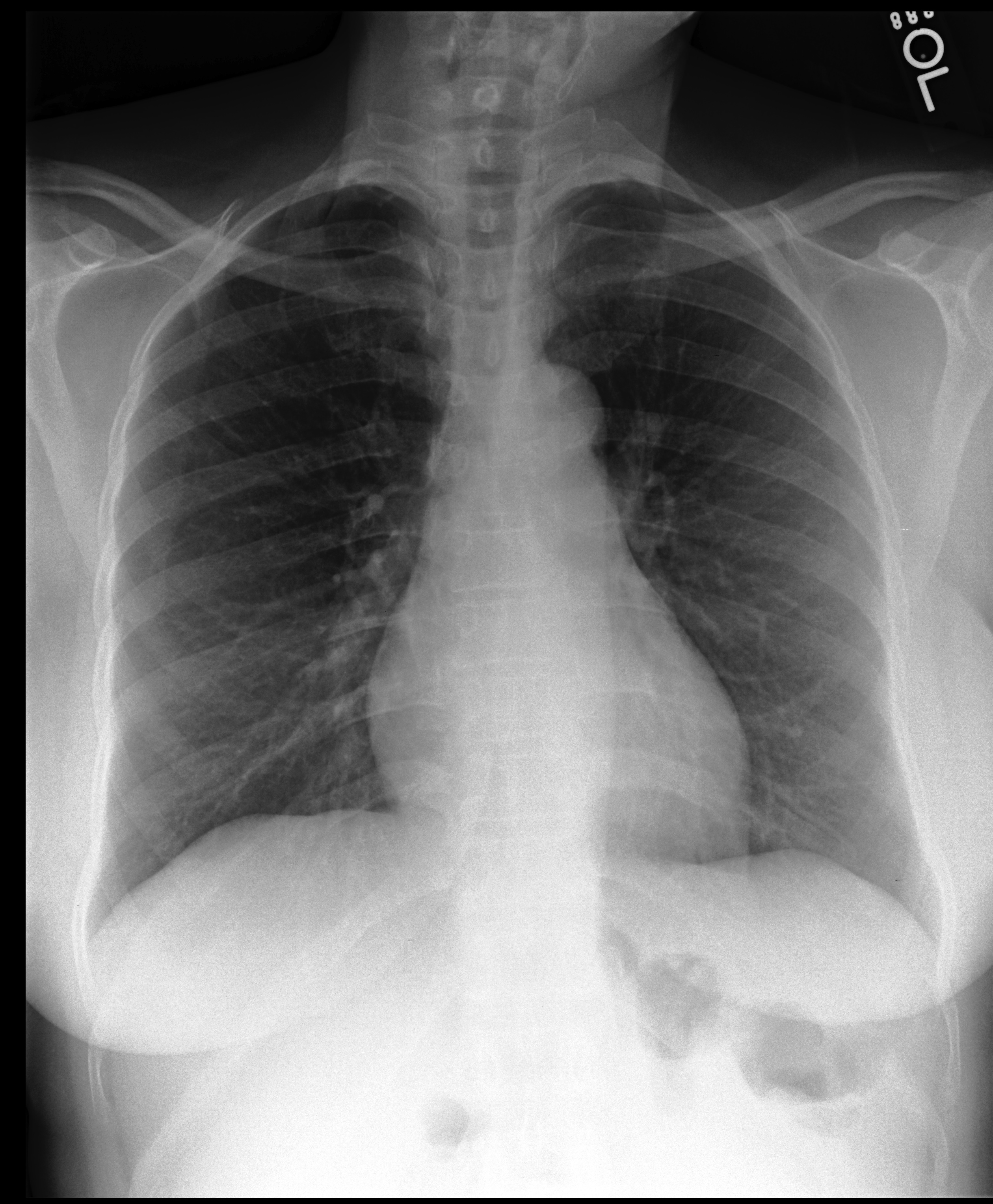}
    \end{subfigure}
    \hspace{0.03\linewidth}
    \begin{subfigure}[b]{0.45\linewidth}
        \centering
        \includegraphics[width=\linewidth]{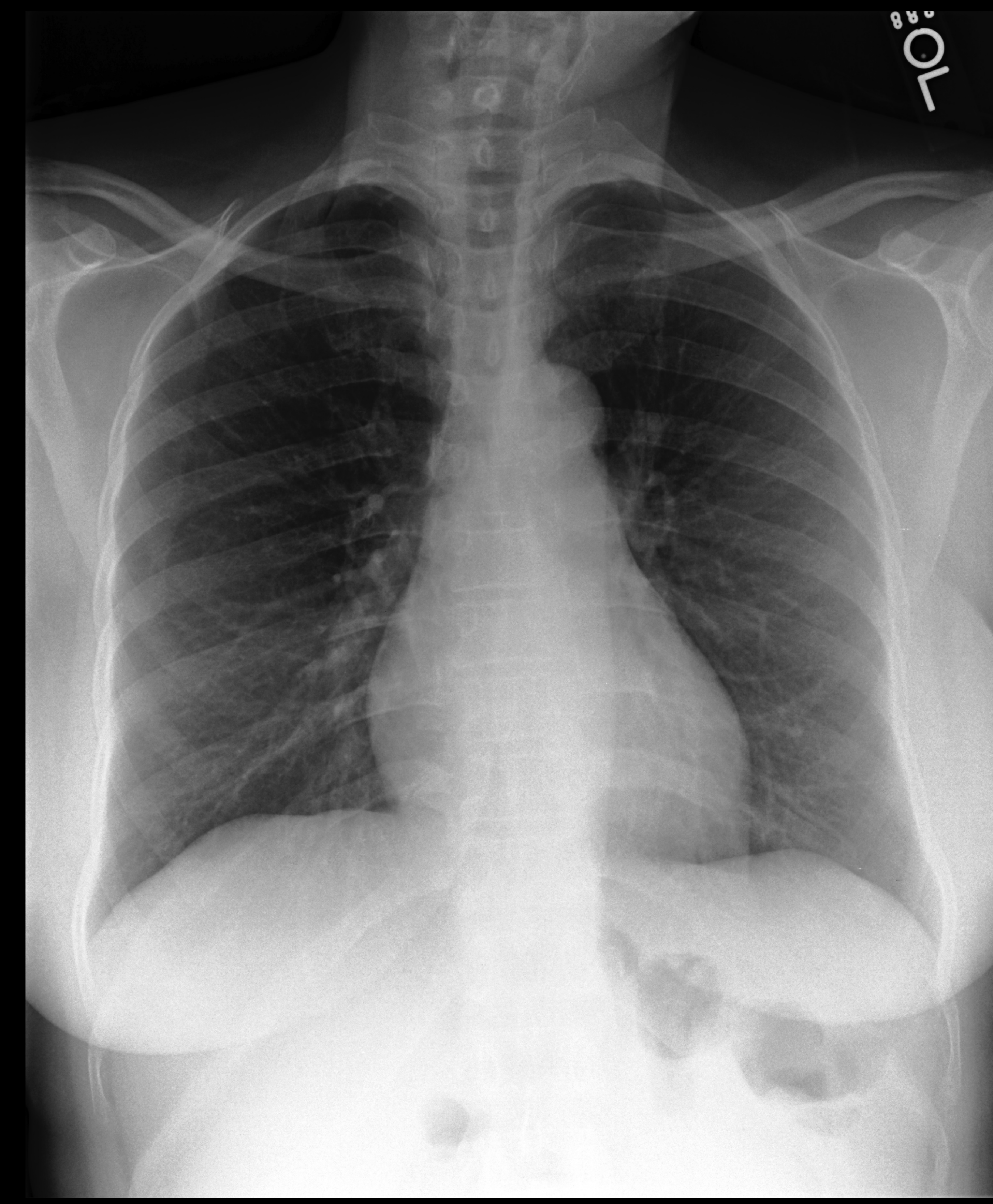}
    \end{subfigure}
    \begin{tcolorbox}
    \textbf{\modelname{}}:
    Findings: Heart size is normal.  The mediastinal and hilar contours are normal. The pulmonary vasculature is normal. Lungs are clear. No pleural effusion or pneumothorax is seen. There are no acute osseous abnormalities.

    Impression: No acute cardiopulmonary abnormality.
    \end{tcolorbox}
    \begin{tcolorbox}
    \textbf{Reference}:
    Findings: The lungs are clear without focal consolidation.  No pleural effusion or pneumothorax is seen.  Cardiac and mediastinal silhouettes are unremarkable.

    Impression: No acute cardiopulmonary process.
    \end{tcolorbox}
    \caption{
        In this example, the model correctly reports that no abnormality is presented.
    }
    \label{fig:more-grg-3}
\end{figure*}

We provide more qualitative examples and analyses in \Cref{fig:more-grg-1,fig:more-grg-2,fig:more-grg-3}.
These examples further demonstrate the performance of \modelname{} and the significance of visual grounding in radiology report generation.

\section{Prompt Templates}

\paragraph{Grounded Reports Construction}

See \Cref{fig:tag-prompt,fig:filter-prompt}.

\paragraph{Datasets Pre-processing}

See \Cref{fig:mimic-cxr-pp1,fig:mimic-cxr-pp2} for MIMIC-CXR, \Cref{fig:ct-rate-preprocess} for CT-RATE, and \Cref{fig:rocov2-preprocess} for ROCOv2.

\paragraph{Evaluation}

Figure~\ref{fig:eval-prompt} depicts the prompt template for evaluating VQA accuracy.




\begin{figure*}[htbp]
    \begin{tcolorbox}
You are an AI assistant with expertise in radiology. Your main task is to meticulously review a provided radiology report and accurately identify the specified anatomical structures and anomaly findings mentioned in the report.

The names of targets to be identified are primarily specified as follows:

- anatomy list (with optional anatomical modifiers): \{'; '.join(anatomy\_list)\}

- anomaly list: \{'; '.join(anomaly\_list)\}

For each phrase identified as a target, convert it to the following format (similar to a hyperlink in Markdown): [<phrase>](<target>), where "<phrase>" denotes the original text of the identified phrase, "<target>" denotes the name of the target provided above that the phrase is identified as.

Below are requirements:

1. Include anatomic modifiers essential for precise localization when highlighting anatomical structures, such as "right", "left", "upper", "lower", "anterior", "posterior", "pulmonary". But you must not include them when they are not modifying any anatomical structures.

2. Exclude any target explicitly stated as absent, negated, or otherwise indicated as not present or uncertain in the findings. For example, nothing should be included in the following negative statements:
  - There is no pleural effusion or pneumothorax
  - No pleural effusion, pneumothorax, or focal consolidation is present.

3. A special case to tag: the enlargement of cardiac silhouette or heart can be tagged as "cardiomegaly".

4. Do not include targets that are too coarse, ambiguous, or amorphous to be spatially localized, such as "free fluid", "chest", "abdomen", "left".

5. The output should be exactly the original text extended with additional tags. Do not alter the input, or generate any additional information.
"
    \end{tcolorbox}
    \caption{
        Prompt template for Llama 3 70B used to identify key phrases.
        Hand-crafted few-shot examples are appended to the prompt.
        Some Python script is presented in the template for simplicity.
    }
    \label{fig:tag-prompt}
\end{figure*}

\begin{figure*}
    \begin{tcolorbox}
You are an AI assistant with expertise in radiology. You will be given with a preliminarily annotated radiology report. In the given report, some of the phrases of anatomical structures and anomaly findings are annotated with the following format: [<phrase>](<target>), where "<phrase>" denotes the original text of the annotated phrase, "<target>" denotes the standard name of the corresponding target.

However, targets that are mentioned to be non-existent in the report text may be wrongly included for annotating.  Therefore, your primary task is to check each annotated entity and its context in the given report, remove the annotation tags of phrases that are indicated as non-existent in the report text. For example, phrases that are described with terms like 'no', 'without', 'absent', 'not detected', 'not observed', 'grossly unremarkable', 'cannot be assessed', or any other negations indicating non-existence. To do the removal, for each annotation of "[<phrase>](<target>)" to be removed, convert it to "<phrase>". On the other hand, annotation tags of targets that are mentioned as being present or observed should still be retained.

Your output should be exactly the same as the original text, except for annotations tags removed for targets that are mentioned to be absent. DO NOT output any additional information, such as your own comments. Also DO NOT add new annotation tags. Even if you find that there is no tags to be removed, the output should be the same as input with all tags kept.
    \end{tcolorbox}
    \caption{
        Prompt template for Llama 3 70B used to filter positive targets.
        Hand-crafted few-shot examples are appended to the prompt.
    }
    \label{fig:filter-prompt}
\end{figure*}

\begin{figure*}
    \begin{tcolorbox}
You are an AI assistant with expertise in radiology. You are given a radiology report. Your task is to process the report and remove contents that is impossible to be inferred solely from a single radiograph.
Specifically, you should:

1. Remove clinical meta information about the imaging planes and techniques and the patient's position, like "AP and lateral views of the chest were provided", "evaluation is limited due to significant patient rotation to the right", "portable chest radiograph", "AP single view of the chest has been obtained with patient in sitting semi-upright position", "frontal images of the chest", "portable AP view of the chest", "on the lateral view", "is identified on both frontal and lateral views".

2. If such contents imply key findings, do paraphrase to retain the key information while performing the removals as requested. For example, "portable chest radiograph shows improved aeration at the right lung base" should be paraphrased to "aeration is seen at the right lung base" and "portable chest radiograph demonstrates a right pneumothorax" should be paraphrased to "a right pneumothorax is seen".

3. Avoid unnecessary removals and paraphrases. Modify the input as little as possible while meeting the above criteria.

Here is the input text for your task:

Input: \{input\}

Your output should be exactly the processed report. Do not output anything else.
    \end{tcolorbox}
    \caption{Prompt templates used to pre-process the MIMIC-CXR dataset in two sequential steps (step 1).}
    \label{fig:mimic-cxr-pp1}
\end{figure*}

\begin{figure*}
    \begin{tcolorbox}
You are an AI assistant with expertise in radiology. You are given a radiology report. Your task is to process the report and remove contents that is impossible to be inferred solely from a single radiograph.
Specifically, you should:

1. Remove comparison with prior examinations and description of interval changes, like "no significant change compared to the prior radiograph", "are similar to prior", "are again noted", "are compared to previous exam from \_\_\_", "since the prior radiograph", "there has been little interval change", "continues to be", "is re-demonstrated", "persistent", "unchanged", "as expected", "stable", "with possible slight decrease in", "perhaps somewhat decreased", "there is increased", "new", "previously", "known".

2. Remove the medical history of the patient and judgements derived purely from it, like "the patient has had prior sternotomy and aortic valve repair", "is consistent with remote history of fracture", "which is compatible with provided clinical history of ILD", "the patient is status post median sternotomy, CABG, vascular stenting", "bilateral pleural catheters have been removed", "consistent with prior granulomatous disease", "the ETT has been removed", "in view of history, a possibility of lymphangitic carcinomatosis also needs to be ruled out".

3. If such contents imply key findings, do paraphrase to retain the key information while performing the removals as requested. For example, "as compared to the prior radiograph performed yesterday morning, there has been slight interval improvement in extent of interstitial pulmonary edema" should be paraphrased to "there is interstitial pulmonary edema", "portable chest radiograph shows improved aeration at the right lung base" should be paraphrased to "there is aeration at the right lung base", "relatively increased opacity projecting over the right lung base is seen" should be paraphrased to "opacity projecting over the right lung base is seen", and "the right lower lobe opacification has decreased substantially" should be paraphrased to "right lower lobe opacification are present".

4. If such contents only describe interval changes relative to prior and whether the abnormalities are currently present cannot be definitely inferred, remove them entirely. For example, "the mediastinal and hilar contours are relatively unchanged", "cardiac and mediastinal silhouettes are stable", "cardiomediastinal silhouette is unchanged" and "no new focal consolidation is seen" should be removed.

5. Avoid unnecessary removals and paraphrases. Modify the input as little as possible while meeting the above criteria.

Here is the input text for your task:

Input: \{input\}

Your output should be exactly the processed report. Do not output anything else.
    \end{tcolorbox}
    \caption{Prompt templates used to pre-process the MIMIC-CXR dataset in two sequential steps (step 2).}
    \label{fig:mimic-cxr-pp2}
\end{figure*}

\begin{figure*}
    \begin{tcolorbox}
You are an AI assistant with expertise in radiology. You are given a radiology report. You should:

1. Remove comparison with prior examinations and description of interval changes, like "prior right rib fractures.", "newly developed", "newly emerged", "stable", "with the patient's previous examinations".

2. Remove the medical history of the patient, like "in the case with a history of perforation during dilatation due to achalasia", "previous pleura in a patient with a history of previous TB", "mentioned in the patient's clinical information may cause these findings".

3. Keep the rest of the report exactly the same without any modification.

Here is the input text for your task:

Input: \{input\}

Your output should be exactly the processed report. Do not output anything else.
    \end{tcolorbox}
    \caption{Prompt template for Llama 3 70B used to pre-process the CT-RATE dataset. Due to the relatively infrequent reference of external information, we only pre-process reports with keywords: "prior", "previous", "new", "stable", "patient" and "history".}
    \label{fig:ct-rate-preprocess}
\end{figure*}

\begin{figure*}
    \begin{tcolorbox}
You are an AI assistant with expertise in radiology. You are given a caption of a radiological image. You should:

1. Remove the patient's personal information, like "a 26-year-old male patient".

2. Remove comparison with prior examinations and description of interval changes, like "comparing to prior studies", "in the previous CT", "previously noticed", "redemonstrated", "unchanged", "new".

3. Remove the medical history of the patient, like "with no previous history of disease", "previous liver surgery".

4. Remove references to figures and cases, like "in Figure 1", "for Case 2", but retain references to arrows.

5. Remove the date of the imaging study, like "taken five days after", "six months postoperative".

6. For the rest of the text that has no content to be removed, keep it exactly the same without any modification.

7. If you find the provided input text does not appear to be a caption of a radiological image, such as it does not mention any radiology-related concepts or terms, then your output should be exactly "The provided input text does not appear to be a caption of a radiological image.".

Here is the input text for your task: 

Input: \{input\}

Your output should be exactly the processed caption, or report that the input text does not appear to be a caption of a radiological image. Do not output anything else, such as other comments.

    \end{tcolorbox}
    \caption{Prompt template for Llama 3 70B used to pre-process the ROCOv2 dataset.}
    \label{fig:rocov2-preprocess}
\end{figure*}

\begin{figure*}
    \begin{tcolorbox}
Your task is to evaluate the correctness of the prediction based on the question and ground truth in a clinical diagnosis scenario.

Question: "\{question\}"

Ground truth: "\{answer\}"

Prediction: "\{prediction\}"

Is the prediction correct? Provide a concise analysis and give an integer score of 0 or 1. Answer in the format "Analysis: ... Score: ...".
    \end{tcolorbox}
    \caption{Prompt template for Llama 3 70B used to evaluate VQA accuracy. We instruct the model to provide analysis befor giving the score to achieve Chain-of-Thought prompting and better explainability.}
    \label{fig:eval-prompt}
\end{figure*}

\end{document}